\documentclass[runningheads]{llncs}
\usepackage[T1]{fontenc}
\usepackage{graphicx}
\usepackage{rotating}
\usepackage{multirow}
\usepackage{multicol}
\usepackage{tabularx}
\usepackage{longtable}
\usepackage{comment}
\usepackage{amsmath,amssymb}
\usepackage[greek,english]{babel}
\usepackage{alphabeta}
\usepackage{bm}
\usepackage{url}
\usepackage{float}
\usepackage[usenames,dvipsnames,table]{xcolor}
\usepackage{booktabs}
\usepackage{mdframed}
\usepackage{graphicx}
\usepackage{caption}
\usepackage{subcaption}
\usepackage{algorithm}
\usepackage{algorithmic}
\usepackage{wrapfig}
\usepackage{caption}
\usepackage[T1]{fontenc}
\usepackage{times}
\usepackage{adjustbox}
\usepackage{pifont}
\usepackage[table]{xcolor}
\usepackage[absolute,overlay]{textpos}
\usepackage{mdwlist}
\usepackage{setspace}
\usepackage[a4paper, twoside=false, left=2.5cm, right=1.5cm, top=3.75cm, bottom=2.5cm, headsep=1.2cm]{geometry} 
\usepackage{csquotes}
\usepackage[twoside]{fancyhdr}
\usepackage{lastpage}
\usepackage{enumerate}
\usepackage[textsize=scriptsize,colorinlistoftodos]{todonotes}
\usepackage{comment}
\usepackage[colorlinks, plainpages=false, hyperindex, pageanchor=true, hyperfootnotes=false]{hyperref}
\usepackage[printonlyused]{acronym}
\usepackage[utf8]{inputenc}

\usepackage{pifont}

\begin{document}
\title{Graph Neural Networks, Deep Reinforcement Learning and Probabilistic Topic Modeling for Strategic Multiagent Settings}
%
%
\author{
Georgios Chalkiadakis\inst{1}
\and
Charilaos Akasiadis\inst{1,2}
\and
Gerasimos Koresis\inst{1}
\and
Stergios Plataniotis\inst{1}
\and
Leonidas Bakopoulos\inst{1}
}
\authorrunning{G. Chalkiadakis et al.}
%
\institute{School of Electrical and Computer Engineering, Technical University of Crete, Chania, Greece
\email{\{gchalkiadakis,cakasiadis,gkoresis,splataniotis,lbakopoulos\}@tuc.gr}\\
\url{https://www.intelligence.tuc.gr/gogs/} \and
Institute of Informatics \& Telecommunications, NCSR `Demokritos', Agia Paraskevi, Greece \\
}
\maketitle              
\begin{abstract}
This paper 
provides a
comprehensive review of mainly Graph Neural Networks, Deep Reinforcement Learning, and Probabilistic Topic Modeling methods with a focus on their potential incorporation in
strategic multiagent settings. We draw interest in {\em (i)} Machine Learning methods currently utilized for uncovering unknown model structures adaptable to the task of strategic opponent modeling, and {\em (ii)} the integration of these methods with Game Theoretic concepts that avoid relying on assumptions often invalid in real-world scenarios, such as the Common Prior Assumption (CPA) and the Self-Interest Hypothesis (SIH). 
We analyze the ability to handle uncertainty and heterogeneity, two characteristics that are very common in real-world application cases, as well as scalability.
As a potential answer to effectively modeling relationships and interactions in multiagent settings, we champion the use of {\em Graph Neural Networks (GNN)}. Such approaches are designed to operate upon graph-structured data, and have been shown to be a very powerful tool for performing tasks such as node classification and link prediction. Past related work indicates that GNNs are useful in cases of multiagent policy learning, and is a good step for overcoming restricting assumptions such as the CPA and SIH.
Next, we review the domain of {\em reinforcement learning (RL)}, and in particular that of {\em multiagent deep reinforcement learning (MADRL)}. Single-agent deep RL (DRL) has been widely used for decision making in demanding game settings. Its application in multiagent settings though is hindered due to, e.g., varying relationships between agents, and non-stationarity of the environment. 
We describe existing relevant game theoretic solution concepts, and consider properties such as fairness and stability. Our review comes complete with a note on the literature that utilizes {\em probabilistic topic modeling (PTM)} in domains other than that of document analysis and classification. The ability of probabilistic topic models to estimate unknown underlying distributions can help with tackling heterogeneity and unknown agent beliefs.
Finally, we identify certain open challenges---specifically, the need to {\em (i)} fit non-stationary environments, {\em (ii)} balance the degrees of stability and adaptation, {\em (iii)} tackle uncertainty and heterogeneity, {\em (iv)} guarantee scalability and solution tractability.

\keywords{Multiagent Deep Reinforcement Learning  \and Common Prior Assumption \and Self-Interest Hypothesis.}
\end{abstract}

\section{Introduction}

In recent years, 
{\em multiagent systems (MAS)}~\cite{wooldridge2009introduction} have gathered increasing attention due to their applicability in various domains such as for microeconomics, for trading in online markets, for optimizing automated decision-making in cooperative and non-cooperative settings, and in general for leading to significant advancements towards the solution of complex real-world problems~\cite{demazeau2010trends,10.1007/978-3-540-75144-1_12,xie2017multi}. The complexity of these systems arises from the interactions between agents~\cite{lutzenberger2016multi}, the uncertainty in the environment~\cite{Nguyen_Kumar_Lau_2017,Varakantham_Cheng_Gordon_Ahmed_2021}, and the need for coordination to achieve common or even conflicting goals~\cite{wooldridge2009introduction}. Traditional approaches often struggle to handle the dynamic nature and non-linear interactions inherent in multiagent (MA) settings.

MAS involve multiple agents interacting with each other in dynamic and often uncertain environments, requiring sophisticated approaches to coordination, decision-making, and learning. 
In such cases, the capability to accurately model opponents---or other players in general---is a crucial component~\cite{10.1007/3-540-60923-7_18,NEURIPS2022_b528459c}.
Agents in MAS often need to {\em adapt their strategies} based on the actions and behaviors of other agents, thus it is desirable to predict the likely actions of their counterparts by learning from past interactions, enabling them to make informed decisions and anticipate possible outcomes.
MA environments are typically {\em dynamic}, with agents continuously updating their strategies and responding to changes in the environment. {\em Opponent modeling}~\cite{DBLP:journals/ai/AlbrechtS18} provides a mechanism for agents to track and to tackle these changes, helping them maintain effectiveness in dynamic and uncertain settings. 
Moreover, understanding the {\em strengths and weaknesses of opponents} is essential for devising effective strategies so that agents can identify patterns, exploit vulnerabilities, and capitalize on the mistakes or suboptimal decisions made by their counterparts, thereby gaining a competitive advantage.

In scenarios where agents need to {\em collaborate or negotiate} with each other, opponent modeling facilitates better understanding of the intentions, preferences, and behaviors of other agents. This understanding enables improved communication, coordination, and negotiation strategies, leading to more efficient and effective collaboration among agents.
In parallel, {\em uncertainty} is inherent in multiagent systems due to incomplete information, stochasticity, and partial observability. Opponent modeling helps reduce uncertainty by providing agents with probabilistic estimates of opponent behaviors, thereby enabling more robust decision-making in uncertain environments.
Finally, by observing and modeling the behaviors of opponents over time, agents can {\em learn from experience} and improve their strategies iteratively. This means that agents should be capable of updating their models dynamically based on new observations, feedback, and interactions, leading to continuous improvement and adaptation~\cite{weyns2009self}.

To this end, a series of methods have been proposed that aim to solve the problem of learning the beliefs of one's opponent(s)~\cite{nyarko2002experimental,pmlr-v80-raileanu18a,zhai2023dynamic}, as well as the internal model structures of other agents~\cite{858456,nashed2022survey}. According to such paradigms
agents may observe the other agents' past actions and payoffs (or, interestingly, observations and actions of the modelled agents may be replicated relying only on local
information of the controlled agent~\cite{papoudakisCA2021nips}); but the underlying belief models are private and must be inferred in some way. 
To achieve this, usually opponent agents are considered as separate entities that are independent from the environment, and a number of ``exploratory'' interactions are required so as to gather the data required to build the corresponding models.
Roughly, this process comprises a method for collecting data, a learning algorithm, and the decision regarding the target resolution, i.e. to learn the type of each agent, to predict the next actions of individuals, or to estimate the collective strategy of a team of agents.
However, assuming that entities are independent from the environment is realistic only in very simple settings.

As a possible solution, 
we examine the adoption of {\em Graph Neural Networks (GNNs)}~\cite{4700287}, which have demonstrated effectiveness in diverse domains such as molecular structure learning, protein function prediction, disease classification, and speech recognition. Notably, recent applications have extended to predicting outcomes in turn-based strategy games and executing specific real-time strategy games using multi-agent deep reinforcement learning algorithms. However, to our knowledge, GNNs have yet to be leveraged for modeling intricate or uncertain strategic interaction domains or aiding in equilibrium learning within such contexts.

Now, a large portion of existing opponent modeling methods relies on single-agent reinforcement learning (RL) with the aim to either maximize individual utility or the overall social welfare. 
However, the problem setting when transiting from single agent to multiagent learning environments changes substantially, as the latter poses more challenges. Specifically, theoretical convergence guarantees are hindered due to non-stationarity, incomplete information, and the fact that different agent actions change the states of the environment in ways that are hard to estimate. Also, it is not certain that the Markov assumption continues to hold.
Taking these into account, more types of multiagent learning methods have been applied, including unsupervised learning techniques, e.g. that employ PTM, and other methods that model the whole agent population policies, e.g. co-evolutionary policy learning techniques, swarm intelligence, or learn the interaction protocols in effect, e.g. via adaptive mechanism design~\cite{10.1007/978-3-030-01713-2_1}.

The particular challenges that arise in strategic MA settings have been a subject of study by past research. One important issue is the exponential growth of the computational complexity of classic RL algorithms when the state-action pairs come in large numbers. In MAS this is further worsened as the number of agents is an additional factor that adds to this exponential growth~\cite{Busoniu2010}. 
Another issue is that of the resolution of the different agent utility and payoffs, since these are typically correlated and their independent maximization is often not realizable. Moreover, the various agents learn simultaneously as they act in the environment leading to the problem of the best policy changing dynamically and not being a stationary point. 
Adding to this, the exploration-exploitation trade-off becomes even more difficult to obtain in such multiagent settings and it is more probable to destabilize the learning process~\cite{Busoniu2010}.

One can categorize multiagent reinforcement learning (MARL) into three types~\cite{tuyls2012multiagent}: {\em multiplied learning}, {\em divided learning}, and {\em interactive learning}. 
In the first case, the learning task is performed independently by each agent and the interactions among them do not affect the way they learn. The resulting strategies can be considered a generalization that can be applied in any setting. In the divided learning case, each agent is assigned particular tasks and the result can be combined to achieve a common goal. Contrary to the previous case, the strategies that each agent learns correspond to different areas or domains, and, if combined, can lead to social welfare maximization. 
Finally, interactive learning techniques can be considered more sophisticated than the other categories, since
this type of learning aims to strike a balance between the two previous extremes, generalization and specialization. In this case the learning tasks are not known a priori, and agents need to explore and negotiate a successful learning path. Agents regulate the joint learning procedure and then synthesize the various perspectives and conflicts that the learning results may pose.

Furthermore, {\em deep reinforcement learning (DRL)} has emerged in the recent years as a powerful tool for decision making in
strategic games, such as Go~\cite{mnih2015human,silver2017mastering}. DRL essentially employs the power of modern deep networks to perform
function approximation in reinforcement learning environments~\cite{sutton2018reinforcement}. However, DRL methods have so far been
used mainly in two- and not multi- player strategic games; and typically require a large number of learning
experiences to reach a satisfactory level of performance.

Overall, in MARL, agents take part in a stochastic game with unknown reward and/or transition models~\cite{littman1994markov,littman2001value,shapley1953stochastic}, and must learn to (inter)act optimally in the presence of the other agents doing the same, and also considering the incompleteness of information, the vast sizes of state spaces, and the distribution of rewards.
The agents try to maximize their expected discounted rewards; 
while their optimal behavior can be defined in terms of game theoretic solution concepts (such as some variant of the well-known Nash equilibrium)~\cite{shoham2007if}.
However, note that demanding convergence of the learning process to some sort of equilibria should not necessarily be seen as a requirement or even as a ``nice-to-have'' property in real-world multiagent systems~\cite{stone2007multiagent}.
In general, MARL should not be considered just as the multiagent extension of the reinforcement learning problem. Rather, in most if not all cases, a MARL method should be explicitly taking other agents into account; and, at the same time, one could be using tools provided by, e.g., game theory to reason about others and thus improve their own performance.

Notwithstanding it being widely employed in modern MAS and AI research~\cite{chalkiadakis2022computational,shoham2008multiagent}, modern algorithmic game theory (AGT)~\cite{roughgarden2010algorithmic} has been criticized for offering, more often than not, negative computational complexity results. 
Moreover, key game-theoretic solution concepts, such as the celebrated NE, have been criticized for {\em (a)} being incapable of characterizing the nature of most real-world equilibria situations as argued~\cite{daskalakis2008complexity}; and for {\em (b)} constituting a controversial ``target’’ or ``holy grail’’ for rational behavior in multiagent learning environments~\cite{chalkiadakis2003coordination,shoham2007if,stone2007multiagent}. 
These criticisms are based mainly on the high computational complexity of NE and related concepts, and to the often unrealistic full-rationality, or other simplifying assumptions adopted in GT. Indeed, even if one would be able to compute equilibria within a reasonable time-frame, the rationality of the equilibrium-prescribed behaviour would often be questionable, given the assumptions usually present in existing equilibrium notions. Therefore, our view is that more steps need to be taken to remove restrictive GT assumptions and practices, and integrate GT further with modern AI and MAS research, to practical benefits.

One specific assumption that is central to the usually studied equilibrium notions is the CPA---namely, the assumption that the beliefs of the various agents are posteriors, obtained by conditioning a common prior on private information received by the individuals. The common prior assumption is a substantive assumption, and alongside certain common knowledge assumptions, it plays an important role in game theory and economics: specifically, it is central to the definition of Bayesian (or ``Bayes-Nash'') equilibria~\cite{shoham2008multiagent}, and is used to restrict attention to correlated equilibria, and (when beliefs are uncorrelated) to NE. As such, the CPA has provided the main decision-theoretic justification of such concepts (see, e.g.,~\cite{aumann1987correlated}). Convenient as this assumption might be, prior information is not, in fact, common in most real-world environments. Moreover, there exist problem settings where explicitly modeling prior beliefs as heterogeneous can improve our understanding and tackling of such problems~\cite{morris1995common}.

Nevertheless, arguments in favor of the CPA’s plausibility exist, such as that if people have always been fed the same information, then it is not unreasonable to assume a common prior~\cite{aumann1987correlated}. 
However, the CPA’s implication that there exists one ex ante, prior stage during which all agents share the same beliefs regarding each other’s beliefs, including beliefs regarding their own actions (and regarding how these beliefs are going to be revised), is, arguably, highly improbable~\cite{gul1998comment,morris1995common}.
Moreover, the assumption of a commonly shared prior during some prior stage of deliberations, runs counter to the Bayesian view of statistics, and of interpreting information and beliefs~\cite{jaynes2003probability}, since it amounts to asserting that at some moment in time everyone must have had identical beliefs.

To the best of our knowledge, all criticisms against the CPA and all attempts to replace its problematic conceptual elements, have so far been provided from a mainly philosophical/theoretical viewpoint. To date, there has been no convincing attempt to override CPA in practice, through explicitly considering a truly dynamic stage through which agents form their initial beliefs. Moreover, one of the most compelling arguments in favour of the CPA is that it considerably simplifies analysis~\cite{morris1995common}. However, we argue  this need not be the case: the use of modern Bayesian techniques can ease the analysis, even under heterogeneous priors.

Moreover, another widely used assumption which has been questioned is the SIH according
to which agents’ objective is the maximization of their personal payoff. Namely, experiments conducted
primarily in the 1980s and 1990s demonstrated that people are not in general self-interested and, as a result,
many scientists proposed models which attempt to explain this phenomenon by introducing in agents’ decision-
making processes the notions of reciprocity, fairness and altruism to name a few~\cite{fehr2006economics,sobel2005interdependent}. Nevertheless,
economists have tried to provide a justification for deviations from solution concepts such as the NE, and never
tried to adopt a bottom-up approach, e.g. from agent characteristics to agent interactions. 
We anticipate that 
by using machine learning techniques in order to reveal the structure of an agent’s thinking and reasoning this can be achieved.

In order to remove the common prior assumption in practice, we need to explicitly model the stage during which agents acquire and form prior beliefs that are potentially entirely private and heterogeneous. As a result, the need for tools that will allow the automatic emergence of the unknown modes of behavior is required. There are several---e.g., Bayesian---tools that can be used to model such prior beliefs.

This paper is further structured as follows.
In Section~\ref{sec:GNN} we put our focus on GNNs, which are widely used for analyzing and modeling complex relationships inherent in multiagent systems. By treating agents and their interactions as nodes and edges in a graph, GNNs can capture both local and global dependencies, enabling effective representation learning and reasoning. Several studies have demonstrated the efficacy of GNNs in tasks such as graph-based state estimation, multiagent coordination, and decentralized decision-making.

Next, in Section~\ref{sec:DRL} we analyze the most recent and effective RL algorithms, with a focus also on MA applications.
Deep reinforcement learning techniques have shown remarkable success in learning optimal policies for individual agents in complex environments. When applied to multiagent settings, DRL algorithms enable agents to learn strategies through interaction with the environment and other agents.

In Section~\ref{sec:equil} we present the necessary background from the field of GT that can be used to set the required basis for more advanced algorithms and solutions.

In Section~\ref{sec:PTM} we describe probabilistic topic models that operate under the assumption that each document in a corpus is generated from a mixture of latent topics, where a topic is characterized by a distribution over words. Probabilistic topic models aim to infer these latent topics and their distributions from the observed word occurrences in the documents. 

In Section~\ref{sec:challenges} we summarize the open challenges and highlight promising research avenues. 
Finally, in Section~\ref{sec:conclusions} we conclude.

\section{Graph Neural Networks}
\label{sec:GNN}

GNNs is a class of machine learning models designed to operate on graph-structured data. 
Unlike traditional {\em neural networks (NNs)} that process grid-structured data like images or sequences, 
GNNs are able to effectively handle complex relationships and dependencies inherent in graph structured data~\cite{GNN_review}.
Formally, a graph is denoted as \( G = (V, E)\), where $V$ represents the set of nodes (vertices) and $E$ the set of edges (connections) between nodes.
In the context of GNN, a graph consists of entities which are represented as nodes and are connected by edges which correspond to relationships or interactions among them. 
Such structures can be found in various domains, for example in social networks, recommendation systems \cite{recom_sys}, biological networks \cite{4700287}, and more. 
GNNs leverage this structure to be able to analyze graph elements \cite{powerful_xu_2019}.

When operating upon graph data, different prediction tasks can be performed, which are categorized into three general types: graph-level, node-level and edge-level \cite{GNN_review}.
In a graph-level task, the objective is to predict a property or attribute for the entire graph as a whole.
For example, in graph classification, the task might involve predicting the category or label of an entire graph based on the its structure and features \cite{graphlvl,4700287}.
In node-level tasks, we want to predict a characteristic or property for each individual node within the graph. Such task is node classification, where each node is given a label or category depending on its features in the graph \cite{nodelvl,4700287}.
Lastly, edge-level tasks focus on predicting properties or the existence of edges within the graph. This can include tasks such as link prediction, where the likelihood of a connection between pairs of nodes is predicted based on their attributes and the graph structure. Also, it can be applied on edge attribute prediction, where properties of existing edges are predicted, such as edge weights or types \cite{edgelvl}.

GNNs are able to capture the process of message exchange between nodes. At each step, nodes exchange information (messages) with their neighbors. 
This process allows nodes to aggregate information from their local neighborhood, 
enabling them to make informed decisions about their own properties or the properties of neighboring nodes \cite{GNN_survey}.
One of the most prominent features of GNNs, is their ability to incorporate contextual information from neighboring nodes \cite{GNN_review}. For example, in a node classification task, each node can aggregate information from its neighboring nodes to refine its own---and possibly not so far known---classification label.

\subsection{GNN variations}
While GNNs offer powerful capabilities for learning from graph-structured data, researchers have developed a number of variants to address specific challenges that exist in the different application domains. Some notable variants of GNNs include GCNs \cite{GCN}, GATs \cite{gat} and GraphSAGE \cite{graphsage}.

GCNs 
generalize the concept of convolutional neural networks to graphs, 
enabling effective information propagation and feature learning across graph nodes. 
They have been successfully applied in tasks such as node classification and graph classification \cite{GCN}.
At the heart of GCNs lies the notion of graph convolution, which enables nodes to aggregate and propagate information from their neighbors. Mathematically, the graph convolution operation in GCN can be expressed as follows:
\begin{equation}
    H^{(l + 1)} = \sigma (\widetilde{D}^{- \frac{1}{2}} \widetilde{A} \widetilde{D}^{- \frac{1}{2}} H^{(l)} W^{(l)})
\end{equation}
where $H^{(l)}$ represents the node feature matrix at layer $l$ for every single node, $\widetilde{A}$ is the adjacency matrix of the graph with self-loops added, $\widetilde{D}$ is the corresponding degree matrix, which is a diagonal matrix with the summation row-wise of the amount of connections per node, $W^{(l)}$ is the learnable weight matrix for layer $l$ used to linearly transform the node features, and $\sigma$ denotes the activation function such as ReLU \cite{GCN}, applied element-wise to the result of the linear transformation.

\begin{figure}[H]
    \centering
    \includegraphics[scale = 0.3]{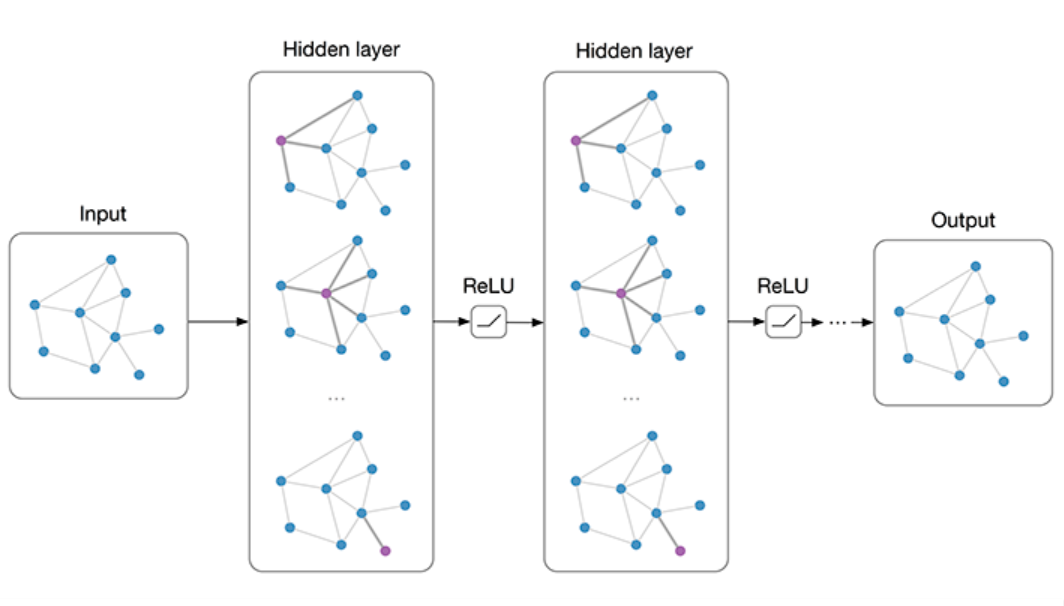}
    \caption{GCN architecture \cite{GCN}}
    \label{fig:gcn}
\end{figure}

The linear transformation $H^{(l)} W^{(l)}$ applies a learned weight matrix $W^{(l)}$ to the node features, enabling the network to learn representations that are most relevant for the task at hand.
The adjacency matrix $\widetilde{A}$ and degree matrix $\widetilde{D}$ encode the graph structure. By multiplying  $\widetilde{A}$ with $\widetilde{D}^{- \frac{1}{2}}$ on both sides, we are able to normalize the adjacency matrix, which aids in stabilizing the training process and improves the model's performance.
The resulting matrix is multiplied with the node features matrix $H^{(l)}$, effectively aggregating information from neighboring nodes. This step captures the graph structure and enables nodes to exchange information with their neighbors.
Lastly, the activation function $\sigma$ is applied element-wise to introduce non-linearity to the model, enabling it to learn complex relationships in the data.
The process described above, is visually represented in Figure~\ref{fig:gcn}.

To illustrate the application of GCNs, consider the task of node classification in a social network. Each node in the graph represents a user, and the edges denote relationships between users (e.g., friendship). By leveraging GCNs, we can learn representations of users based on their interactions and predict user attributes or labels. For example, we can predict the demographic characteristics of users (e.g., age, gender) based on their network connections and activity patterns.
Furthermore, GCNs are successfully applied for graph classification tasks, where the goal is to classify entire graphs based on their structural properties. For instance, in molecular graph analysis each graph represents a chemical compound, and the nodes and edges correspond to atoms and bonds, respectively. GCNs can effectively capture the molecular structure and predict properties of compounds, such as toxicity or drug activity \cite{GNN_review}.
In summary, GCNs provide a powerful framework for learning from graph-structured data, enabling tasks such as node classification and graph classification. By extending convolutional operations to graphs, GCNs facilitate information propagation and feature learning across nodes, making them suitable for a wide range of applications in domains such as social networks, biology, and recommendation systems.

Another noteworthy extension of GNNs are GATs \cite{gat,GNN_review}, which emerge as a neural network architecture for processing graph-structured data.
Whether it is for modeling social networks, molecular structures, or language syntax, GATs leverage attention mechanisms to weight the importance of the different nodes and edges in a graph.
GATs introduce attention mechanisms into the message passing process, allowing nodes to dynamically weigh the importance of information received from their neighbors \cite{attention_vaswani_2017}. By assigning attention coefficients to edges, GATs can capture complex relationships and dependencies in graph data more effectively \cite{gat}.

\begin{figure}[H]
    \centering
    \includegraphics[scale = 0.4]{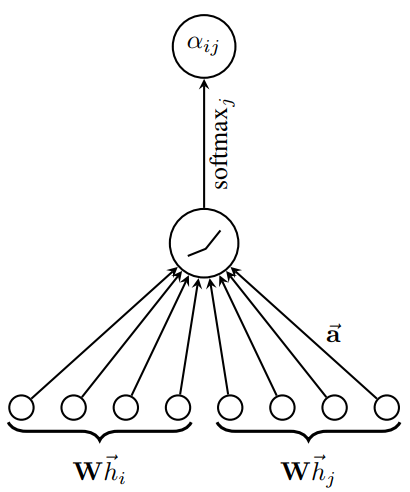}
    \caption{Computation of attention coefficients \cite{gat}}
    \label{fig:att_coeff}
\end{figure}

In a GAT, each edge connecting two nodes is associated with an attention score. 
These scores, denoted as $\alpha_{ij}$ for the edge between node $i$ and node $j$, are computed based on a learned weight matrix $W$, which is trained alongside the rest of the neural network. 
The attention scores $\alpha_{ij}$ reflect the importance or relevance of node $j$ to node $i$ when passing messages \cite{MA_GAT,gat}.
The computation of attention score is typically performed using a form of attention mechanism, such as the softmax function, which ensures that the attention scores sum to one across all neighboring nodes:
\begin{equation}
    \alpha_{ij} = softmax(LeakyReLU(a^T[Wh_i || Wh_j]))
\end{equation}
Here, $h_i$ and $h_j$ are the feature representations of nodes $i$ and $j$, respectively, and $a$ is a learnable attention parameter vector \cite{gat}, the combination of the mentioned elements is also presented in Figure~\ref{fig:att_coeff}.

This attention mechanism allows to concentrate on the most relevant information, thus facilitating more informed and accurate predictions. 
The versatility of GATs is underscored by their application in diverse tasks such as node classification, link prediction, and graph generation~\cite{GNN_review}.
GATs address the limitations of traditional GCNs by enabling nodes to adaptively focus on influential neighbors. This attention mechanism enhances context-aware feature propagation and precise predictions within graphs.
They can also be applied in multi-agent settings, due to the inherent ability of GATs to capture intricate dependencies within graph structures. This allows them to capture evolving relationships between the nodes \cite{MA_GAT}.\newline

To avoid confusion, it is important to highlight some of the differences between GCNs and GATs.
GCNs represent a class of neural networks tailored for graph structures, employing convolutional operations to relay information among neighboring nodes within a graph. This mechanism, allows GCNs to extract informative features representing the graph's topology, which is useful in tasks such as node classification and link prediction \cite{GCN,GNN_review}.
In contrast, GATs use attention mechanisms to dynamically assign importance to adjacent nodes during information propagation. This flexibility enables GATs to learn varying degrees of significance among neighbors, offering a more nuanced approach compared to the fixed weighting employed by GCNs. Notably, GATs have exhibited superior performance over GCNs across various benchmark datasets, particularly in scenarios characterized by sparse graphs or tasks requiring the learning of extensive dependencies \cite{gat,GNN_survey}.

\begin{figure}[H]
    \centering
    \includegraphics[scale = 0.4]{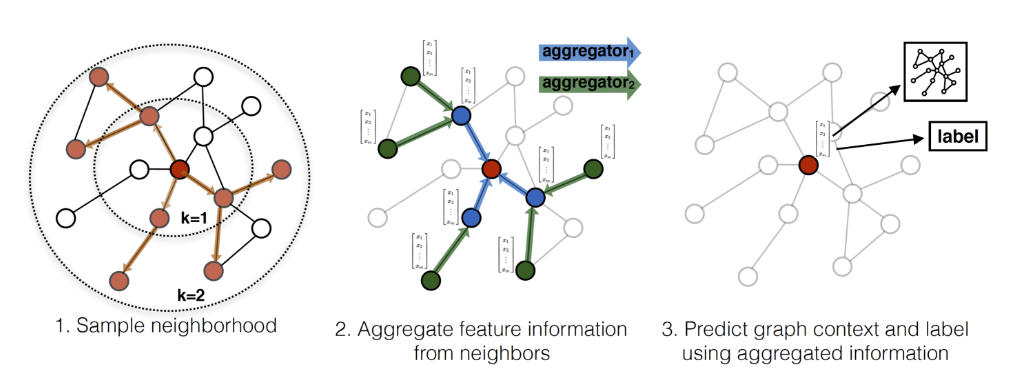}
    \caption{GraphSAGE architecture \cite{graphsage}}
    \label{fig:graphsage}
\end{figure}

GraphSAGE (Graph Sample and Aggregation) is a framework for inductive representation learning on large graphs. It samples and aggregates information from node neighborhoods, enabling scalable and efficient learning on graphs with millions or even billions of nodes \cite{graphsage}. 
Existing embedding methods focus on fixed graphs and use matrix-factorization for each node, lacking the ability to generalize to unseen data. In contrast, GraphSAGE uses the node features in order to develop an embedding function applicable to nodes not present in training. 
Instead of treating each node in isolation, GraphSAGE considers the collective information from a node's neighbors. For example, in a social network, we are interested in learning more about a certain node, in order to get the full picture we also check information about its connections and their characteristics. In a similar manner, to understand the context of a node, GraphSAGE aggregates information from the nodes in the neighborhood of our interest.
To achieve this, GraphSAGE trains a set of aggregator functions. The intuition is that at each training iteration these functions learn to combine information from neighboring nodes at different depths, as shown in Figure~\ref{fig:graphsage}. 
The equation encapsulating the process in Figure~\ref{fig:graphsage} can be represented as:
\begin{equation}
    h^{(k)}_v = AGGREGATE_k \big( \{h^{(k - 1)}_u, \forall u \in N(v)\} \big)
\end{equation}

The aggregated representation of node $v$ at layer $k$ is represented by the term $h^{(k)}_v$ and captures the information from the neighboring nodes up to layer $k$. Respectively, $h^{(k - 1)}_v$ represents the feature vector of a neighboring node $u$ at the previous layer $k - 1$.
The set of neighboring nodes of node $v$ is denoted by $N(v)$.
The component $AGGREGATE_k$ is the aggregation function used to aggregate information from the neighboring nodes. It operates on the feature vectors of neighboring nodes $h^{(k - 1)}_v$ and produces the aggregated representation $h^{(k)}_v$ for node $v$.
For each $v$, we gather the feature vectors $h^{(k - 1)}_v$ of all its neighboring nodes $u$ at the previous layer $k - 1$.
These feature vectors are then aggregated using $AGGREGATE_k$, which computes a summary representation of the neighboring nodes' features, capturing the local neighborhood information of node $v$.
The result $h^{(k)}_v$ captures the information for the neighborhood of node $v$ up to layer $k$, which is then used as input for the next layer.
One of GraphSAGE's prominent features is its ability to be employed in different types of graphs, apart from social networks. It can also be used, for example, to understand the connections between different proteins in a biological network \cite{PPI}. GraphSAGE's performance is mainly tested in node-classification tasks, such as citation networks, where it is able e.g., to predict the topic of a paper by observing only its citations and its abstract \cite{graphsage}.

Having presented the three most important GNN extensions, we now provide an overview of their similarities and differences.
All three variations operate on graph-structured data, leveraging information from a node's neighborhood to compute node embeddings. The goal is to learn meaningful representations for nodes in a graph. Such representations encapsulate the structure and features of nodes, aiding in node classification and link prediction tasks.
The three variants differentiate from one another in the way they aggregate information from a node's neighborhood. In the case of GraphSAGE, fixed-depth sampling and aggregation functions are used. GCNs use convolutional operations (usually a weighted sum or average) and GATs employ attention mechanisms to assign different weights to neighboring nodes dynamically based on their relevance to the target node.
Furthermore, they differ also in how flexible they are in aggregation. GCN is the least flexible due to using fixed aggregation function like mean or max pooling, which can limit its adaptability to different types of graphs or tasks. GraphSAGE and GATs on the other hand, both offer flexibility in aggregation techniques, enabling users to their environment's characteristics and specific requirements.
Lastly, scalability is another component on which these models are different. GraphSAGE is designed to handle large graphs with millions and billions of nodes for inductive learning and it is known for its scalability and efficiency in such scenarios. This however is not the case for GCNs and GATs, since both face scalability challenges in large-scale graphs for having to aggregate all of the graph's neighborhoods for their training.

There are also GNN variants that operate on more complex graphs, such as on multigraphs~\cite{harary1969graph} or hypergraphs~\cite{berge1973graphs,christianos2017efficient}.
For instance, hypergraphs can come in handy for representing settings where communities of nodes share similar characteristics and can be thus seen as belonging in a hyperedge.
How to design and train GNNs on such complex graphs is an active area of research~\cite{DBLP:conf/nips/YadatiNYNLT19,zhong2023hierarchical}.

Another important aspect is the underlying uncertainty in some cases of graph structures. Most of the examined methods are not able to capture uncertainty in case this is inherent in some problem settings. 
An approach that is able to handle such issues though is Bayesian Graph Convolutional Neural Networks~\cite{zhang2019bayesian}.
In Bayesian Neural Networks, the internal network weights are considered as random variables, a fact that renders the neural NN's output to be a random variable as well.
Accordingly, the prediction for a new input $x$ is calculated by integrating with respect to the posterior distribution of the weights $W$:
$$p(y|x,\mathbf{X},\mathbf{Y})=\int p(y|x,W)p(W|\mathbf{X},\mathbf{Y})dW$$
where $y$ is the NN's output, and $\mathbf{X},\mathbf{Y}$ the training inputs and outputs, respectively.
The $p(W|\mathbf{X},\mathbf{Y})$ can be inferred by expectation propagation, variational inference, and Markov Chain Monte Carlo methods.
Similarly, in the GNN setting, the input graphs are assumed to be instances of random graph ``distributions''.
The ultimate goal of Bayesian GCNNs is to come up with the posterior probability distribution of the outputs of $L$ layers (noted as $\mathbf{Z}$ as below:
$$p(\mathbf{Z}|\mathbf{Y}_{\mathcal{L}},\mathbf{X},\mathcal{G}_{obs}) = \int p(\mathbf{Z}|W,\mathcal{G},\mathbf{X})p(W|\mathbf{Y}_{\mathcal{L}},\mathbf{X},\mathcal{G})p(\mathcal{G}|\lambda)p(\lambda|\mathcal{G}_{obs})dW \mbox{ } d\mathcal{G} \mbox{ } d\lambda$$
where $\mathbf{Y}_{\mathcal{L}}$ are the observed node labels, $\mathcal{G}_{obs}$ the observed graph structure, and $\lambda$ parameters that characterize a family of random graphs.
Since this integral is intractable, approximation methods can be used for its calculation, such as variational inference and Monte Carlo Markov Chains.
Bayesian GNNs have been used in a plethora of domains, such as multiple instance learning~\cite{pal2022bag}, network analysis, biochemistry, recommendation systems~\cite{9555949}, as well as in non-parametric graph distribution cases~\cite{pmlr-v124-pal20a}.

In summary, GraphSAGE, GCNs, and GATs share the objective of learning node representations from graph data. They differ in their aggregation mechanisms, flexibility in aggregation techniques, and scalability. GraphSAGE stands out for its efficiency and scalability, while GCNs and GATs offer unique approaches to integrating neighborhood information through convolutional and attention mechanisms, respectively. The choice among these models depends on the specific characteristics of the graph data and the requirements of the task at hand. In case uncertainty with respect to the graph structure exists, Bayesian GNNs can be proven valuable, even without assuming particular prior parameters.

\subsection{GNN limitations}
However, as it is to be expected, GNNs also face their own set of challenges and drawbacks. Computational complexity may escalate with the size of the graph, leading to potential scalability challenges when dealing with large datasets.
Overfitting is another risk, since GNNs might adapt too specifically to the training data, potentially leading to poor generalization results for unseen relationships between nodes.

GNNs can become intricate and challenging to interpret, especially when multiple layers and attention mechanisms are incorporated~\cite{attention_vaswani_2017,gat}. 
Additionally, their performance is contingent on hyperparameter settings, including the number of layers, attention mechanisms, and learning rates. 
GNNs also face challenges in scenarios where limited historical data is available for new or rarely encountered coalition formations. 
Furthermore, temporal dynamics is another element that GNNs face trouble with, as GNNs inherently assume a static graphic structure and may struggle to capture evolving relationships over time.
Overall, GNNs can also be applied---though with appropriate calibration---to multi-agent systems, which inherently possess graph-like structures, where agents and their relationships can be represented as nodes and edges in a graph~\cite{GNN_review}.

\subsection{GNN applications in Multiagent settings}
An important work highlighting the use of GNNs in MA environments is that of \cite{MA_GAT}.
The paper proposes a novel game abstraction mechanism based on two-stage attention network called G2ANet, which models the relationship between agents using a complete graph, to detect interactions and their importance.
The paper addresses the problem of policy learning in large-scale multi-agent systems, where the large number of agents and complex game relationships make the learning process challenging.
In order to capture interactions between agents, existing methods use pre-defined rules, which are not suitable for large-scale environments due to the complexity of transforming those interactions into rules.
The integration of G2ANet into GNN-based MARL is able to simplify the learning process as well as improve asymptotic performance in comparison to state-of-the-art algorithms \cite{MAAC,MADDPG,IC3Net}.

The G2ANet game abstraction mechanism and the proposed learning algorithms aim to simplify the learning process and improve performance in multi-agent systems. This mechanism utilizes the two-stage attention network to model the interactions between agents in a multiagent environment.
By representing the relationships between agents as a complete graph and leveraging attention mechanisms, G2ANet can effectively detect and prioritize  interactions among agents.
This abstraction mechanism is able to simplify the learning process through providing a structured representation of game interactions, enabling efficient policy learning in complex multiagent systems.
The methods used focus on modeling the relationships between agents and infer the importance of their interactions based on attention mechanisms, which aligns with the principles of CPA. 
It does not follow the SIH due to the fact that the paper emphasizes cooperative MARL where agents collaborate to achieve common goals.

Another noteworthy employment of GNNs in a multiagent setting is that in the work of \cite{DICG}.
The Deep Implicit Coordination Graph (DICG) architecture is introduced, with the goal of enhancing coordination in multiagent reinforcement learning scenarios. The DICG consists of a dynamic coordination graph inference module accompanied by a GNN module.
In such scenarios, coordination graphs are usually computed based on factorized representations, which leads to challenges in both scalability and adaptability. However, DICG employs attention mechanisms to learn the agent observation-dependent coordination graph with soft weights \cite{DICG}.
In that way, they avoid relying on domain-specific heuristics and are able to represent complex relationships between agents across various multiagent domains.

Through the utilization of self-attention, DICG is able to capture the strength of connections between agents implicitly, gaining a soft-edge coordination graph. In doing so, they avoid using binary weights for the edges, offering a greater expressiveness for agent relationships as well as increasing the flexibility of the approach.
Furthermore, the DICG employs graph convolution in order to integrate information across agents, aided by the gathered soft-edge coordination graph. By this integration, they can use the insights gained about the interaction among agents in various ways. Such as, directly obtaining actions by embedding them as observations or estimating baseline values for the agents to utilize during policy optimization.

In the above context, the approach used in \cite{DICG} follows the principles of the CPA but not the SIH. This is because it focuses on improving the ability of the agents to learn coordination strategies considering collective goals, by inferring dynamic coordination graph structures.
Overall, the DICG architecture offers a flexible framework for multiagent reinforcement learning, leveraging attention mechanisms and graph neural networks to infer coordination structures. It improves coordination efficiency in domains with a large number of agents. However, it has not yet been applied into off-policy learning and its sampling efficiency shows grounds for improvement in the future. Similar to \cite{DICG} are the methods implemented in \cite{GCRL_MA} and \cite{GNN_MARL_DICGLIKE}.

The work of \cite{gatdrl} is a combination of GNNs in MA settings in the context of resource allocation in dense cellular networks. Their approach focuses on the context of network slicing and aims to address the challenge of efficiently allocating resources to different slices, such as devices and applications, while maximizing the network performance and user satisfaction.
In order to tackle this problem, they leverage GATs and MARL. By utilizing the attention mechanisms of GATs, they can effectively capture the interactions between different network entities and organize relevant information. This way, the complex relationships and dependencies among network elements are effectively modeled, including base stations, users and network slices.
The employment of MARL techniques enables collaborative decision-making among multiple agents, such as base stations, in dynamically allocating resources based on network conditions and user demands.
Such a collaborative approach is able to enhance the adaptability in resource management leading to improved network performance as well as resource utilization.
The proposed approach has broad applications in various scenarios where efficient resource management is crucial, including next-generation cellular networks, edge computing environments, and Internet of Things (IoT) networks. However, it is limited by challenges regarding computational complexity, scalability, and generalization to diverse network environments.
Overall, \cite{gatdrl} presents an innovative solution to address resource management challenges in dense cellular networks using GATs and MARL. Further research and experimentation is required to validate its effectiveness and address potential limitations in real-world deployments. Other works in the same context are those of \cite{marl-gq} and \cite{magnet}. 
Interestingly, this paper does not rely on the CPA, since the coordination mechanism implemented with the use of GATs, is to capture the fluctuating service demands on the network. This does not include any priors that must be updated, but instead tries in real-time to achieve dynamic collaboration among the base stations. Additionally, the approach is arguably not based on the SIH, due to the fact that the goal of the approach is the optimization of the resource allocation strategy, which is a collective goal rather than having each base station try to maximize some individual reward.

The study presented in~\cite{marl-autdriving} explores the domain of autonomous vehicle control, utilizing RL alongside latent state estimation and spatiotemporal correlations. The proposed framework integrates latent estimation with relational representation, introducing an innovative strategy to improve policy efficiency in autonomous driving situations.
The paper introduces a reinforcement learning framework that explicitly seeks to infer the hidden states of adjacent drivers by utilizing GNNs, thereby notably enhancing the performance of autonomous driving policies. It offers designations for hidden states and impact transmission structures, consequently improving the interpretability within autonomous driving situations. The approach integrates hidden state inference and relational modeling, with the objective of enhancing both the success rates of navigation and the accuracy of hidden state inference within intricate driving situations. By employing GNNs as in the proposed STGSage model, we can attain a deep comprehension of spatial relationships and temporal advancements among different traffic entities, thereby enhancing decision-making in intricate driving scenarios.
The results exhibit a significant improvement in the rates of successful navigation, especially in intricate T-intersection situations. Through the integration of latent inference and RL, the model effectively tackles the obstacles related to interpretability and efficacy, presenting a hopeful resolution for autonomous driving mechanisms. The unique strategy not only enhances the accuracy of navigation results but also creates a solid basis for future advancements in the field of autonomous driving technology.
Despite its advancements, the study acknowledges certain limitations, notably the complexity introduced by varying numbers of surrounding vehicles in different scenarios. However, the proposed framework showcases a substantial improvement in latent inference accuracy, particularly in T-intersection scenarios, underscoring its potential for real-world application in autonomous driving contexts.

The work of \cite{marl-autdriving} focuses on using RL techniques to enable autonomous vehicles to navigate complex driving scenarios independently, without relying on a shared understanding or coordination among agents. Thus, the approach does not rely on the CPA. However, it does follow the SIH, since each vehicle learns its own policy without direct coordination with other vehicles. Also, each vehicle is focused on optimizing its own behavior to achieve its objectives, such as reaching its destination safely and efficiently.

Koresis {\em et al.} in ~\cite{koresisHGNNCF2025}
propose harnessing the power of Hypergraph Neural Networks (HGNNs) that fit generic hypergraph-structured historical representations of {\em coalitional task executions}, in order to learn the unknown values of coalitional configurations undertaking the tasks. 
Unlike previous works in the literature, here the coalitional value uncertainty is due to uncertainty regarding the values of synergies among the {\em different agent types}: the unknown values of the agent types’ synergies in completing various coalitional tasks are what is learned by the HGNN.
However, the fitted model by itself cannot be used to provide suggestions on which coalitions to form; it can only be queried for the values of given coalition–task configurations. To actually provide coalitional suggestions, the approach relies on informed search approaches that incorporate HGNN output as an indicator of the quality of the proposed coalition configurations.

Finally, the work of~\cite{rahman2021openadhocteamwork} presents a GNN-based approach to learn the models of agents in {\em open} ad-hoc teamwork settings.
In such settings, a learning agent is tasked with  efficiently collaborating
with an unknown group of teammates whose composition may change over time, since agents
of different types can dynamically enter and leave the team.
Specifically, the learner undertakes the task to fit the behavioral models of participating agents and to estimate the optimal policy in an open ad hoc teamwork setting modeled as a stochastic Bayesian game. 
The proposed {\em Graph-based Policy Learning (GPL)}  approach
puts forward an action-value computation
that integrates GNN-learned agent and joint-action
value models to produce action-value estimates.
In some detail, the method utilizes coordination graphs to recognize the other agents' actions effect on the learner, and then GNN to predict the upcoming actions.
Then,~\cite{rahman2023generallearningframeworkopen} extends the aforementioned framework to also account for the more realistic case of {\em partial} state observability---providing new algorithms specifically designed for such partially observable settings, alongside a novel formal model for open ad hoc teamwork.

\begin{table}[H]
\centering
\resizebox{0.85\textwidth}{!}{%
\begin{tabular}{c|ccccccccc|}
\cline{2-10}
 &
  \multicolumn{9}{c|}{Algorithms} \\ \cline{2-10} 
 &
  \multicolumn{1}{c|}{\textbf{GCN } \cite{GCN}} &
  \multicolumn{1}{c|}{\textbf{GAT} \cite{gat}} &
  \multicolumn{1}{c|}{\textbf{GraphSAGE} \cite{graphsage}} &
  \multicolumn{1}{c|}{\textbf{G2ANet} \cite{MA_GAT}} &
  \multicolumn{1}{c|}{\textbf{DICG} \cite{DICG}} &
  \multicolumn{1}{c|}{\textbf{GAT-based MARL \cite{gatdrl}}} &
  \multicolumn{1}{c|}{\textbf{STGSage \cite{marl-autdriving}}}
  &
    \multicolumn{1}{c|}{\textbf{HGNN \cite{koresisHGNNCF2025}}}
  &
  \multicolumn{1}{c|}{\textbf{GPL \cite{rahman2023generallearningframeworkopen}}}
  \\ \hline
\multicolumn{1}{|c|}{\textbf{Spatial Domain}} &
  \multicolumn{1}{c|}{\ding{55}} &
  \multicolumn{1}{c|}{\ding{51}} &
  \multicolumn{1}{c|}{\ding{51}} &
  \multicolumn{1}{c|}{\ding{51}} &
  \multicolumn{1}{c|}{\ding{55}} &
  \multicolumn{1}{c|}{\ding{51}} &
  \multicolumn{1}{c|}{\ding{51}} &
  \multicolumn{1}{c|}{\ding{51}} &
  \multicolumn{1}{c|}{\ding{51}}
  \\ \hline
\multicolumn{1}{|c|}{\textbf{Spectral Domain}} &
  \multicolumn{1}{c|}{\ding{51}} &
  \multicolumn{1}{c|}{\ding{55}} &
  \multicolumn{1}{c|}{\ding{55}} &
  \multicolumn{1}{c|}{\ding{55}} &
  \multicolumn{1}{c|}{\ding{51}} &
  \multicolumn{1}{c|}{\ding{55}} &
  \multicolumn{1}{c|}{\ding{55}} &
  \multicolumn{1}{c|}{\ding{55}} &
  \multicolumn{1}{c|}{\ding{55}}
  \\ \hline
\multicolumn{1}{|c|}{\textbf{Sequential Aspect}} &
  \multicolumn{1}{c|}{--} &
  \multicolumn{1}{c|}{--} &
  \multicolumn{1}{c|}{--} &
  \multicolumn{1}{c|}{\ding{55}} &
  \multicolumn{1}{c|}{\ding{51}} &
  \multicolumn{1}{c|}{\ding{51}} &
  \multicolumn{1}{c|}{\ding{51}} &
  \multicolumn{1}{c|}{\ding{55}} &
  \multicolumn{1}{c|}{\ding{51}}
  \\ \hline
\multicolumn{1}{|c|}{\textbf{Does not rely on CPA}} &
  \multicolumn{1}{c|}{--} &
  \multicolumn{1}{c|}{--} &
  \multicolumn{1}{c|}{--} &
  \multicolumn{1}{c|}{\ding{55}} &
  \multicolumn{1}{c|}{\ding{55}} &
  \multicolumn{1}{c|}{\ding{51}} &
  \multicolumn{1}{c|}{\ding{51}} &
  \multicolumn{1}{c|}{--} &
  \multicolumn{1}{c|}{\ding{55}}
  \\ \hline
\multicolumn{1}{|c|}{\textbf{Does not rely on SIH}} &
  \multicolumn{1}{c|}{--} &
  \multicolumn{1}{c|}{--} &
  \multicolumn{1}{c|}{--} &
  \multicolumn{1}{c|}{\ding{55}} &
  \multicolumn{1}{c|}{\ding{51}} &
  \multicolumn{1}{c|}{\ding{51}} &
  \multicolumn{1}{c|}{\ding{55}} &
  \multicolumn{1}{c|}{--} &
  \multicolumn{1}{c|}{--} 
  \\ \hline
\end{tabular}%
}
\caption{A taxonomy of the GNN-based algorithms that we cover in our review. The {\em spectral domain} is where graph features are represented in the graph Fourier basis, so ``convolutions'' are defined as filtering by functions of Laplacian eigenvalues. The {\em spatial domain} is where graph convolutions are defined directly on the graph structure by aggregating and transforming information from a node’s local neighborhood (message passing) in the vertex/node domain.}
\label{tab:gnn_taxonomy}
\end{table}

\section{Deep Reinforcement Learning}
\label{sec:DRL}

{\em Reinforcement learning (RL)}~\cite{bertsekas2019reinforcement,Kaelbling1996Reinforcement,Sutton1998Reinforcement} is a specialized area within AI. It equips an agent with the ability to learn via trial and error while repeatedly interacting with an environment; and, importantly, it allows for the optimization of the agent's sequential behavior under uncertainty. In more detail, the agent observes the representation of
the current state of an environment, takes an action that leads to a new state, obtains a reward, observes
the new current state, and so on. The agent’s ultimate goal is to learn a policy---i.e. a mapping of environment states to actions---that maximizes its long-term accumulated reward.

RL is naturally
paired with the Markov Decision Processes (MDPs) framework~\cite{howard1960dynamic,Puterman1994MarkovDP,Wal1985Markov}, which provides a more detailed and rigorous portrayal of an evolving environment and
its dynamics---albeit under certain assumptions, such as stationrity, that might or might not apply in the real world. 
MDPs are employed for modeling
decision making problems, and are an extension of Markov Chains \cite{Joseph2020Markov,Sargent1960Finite}. By definition,
the MDP is a tuple $(\mathcal{S}, \mathcal{A}, \mathcal{T}, \mathcal{R}, \gamma)$ where:

\begin{itemize}
    \item $\mathcal{S}$ is a finite set of states
    \item $\mathcal{A}$ is a finite set of actions
    \item $\mathcal{T}(s, \mathrm{a}, s')$ = $\Pr(s'|s,\mathrm{a})$ is the transition function - a distribution that returns the probability of landing to state $s'$ by doing action $\mathrm{a}$ in state $s$
    \item $\mathcal{R}(s, \mathrm{a},s')$ is the reward function
    \item $\gamma \in [0,1]$ is the discount factor
\end{itemize}

Noticeably, the transition probabilities and the reward function depend only on the current state $s$ and not on all past states (Markov property). 
Another assumption in this formulation, is that the agent has full observability over the environmental state. However, this 
is not always the case since in many problems the agent can only observe certain parts of the  underlying hidden state in effect. To this end, an extension to MDP has been presented, named POMDP \cite{kaelbling1998planning,monahan1982state,Spaan2005Perseus:} that addresses partial observability. 
A POMDP is defined as a tuple $(\mathcal{S}, \mathcal{A}, \mathcal{T}, \mathcal{R}, \Omega, \mathcal{O}, \gamma)$ where:
\begin{itemize}
    \item $\mathcal{S}, \mathcal{A}, \mathcal{T}, \mathcal{R}, \gamma$ are the same as in the MDP formulation
    \item $\Omega$ is a space where all probable observations lie
    \item $\mathcal{O}(s', \mathrm{a}, \omega)=\Pr(\omega|s',a)$ is the observation function that generates the conditional probability for receiving observation $\omega \in \Omega$ after taking action $\mathrm{a}$ and landing in state $s'$
\end{itemize}

A single interaction with the environment goes as follows: The agent lies in a state $s \in \mathcal{S}$, decides to take an action $\mathrm{a} \in \mathcal{A}$ which will result to a new state $s' \in \mathcal{S}$ with probability  $\mathcal{T}(s, \mathrm{a}, s')$ and receives an immediate reward $\mathcal{R}(s, \mathrm{a}, s')$ as well as an observation $\omega \in \Omega$ with probability $\mathrm{O}(s', \mathrm{a}, \omega)$, if partial information is considered. 
The discount factor $\gamma$ also plays an important role in the learning competence of the algorithm, as it determines how much are immediate rewards favored over future rewards. If $\gamma=0$ the 
agent becomes "myopic", taking only the actions that will maximize its 
immediate reward. As $\gamma$ approaches 1, the agent becomes more 
far-sighted, taking actions that maximize its long term reward. Typically, $\gamma$ is kept constant 
throughout the whole training phase and its exact value is selected based on the characteristics of the task at hand.

A solution
to a POMDP consists of a policy $\pi$, which is essentially a mapping from the state space $\mathcal{S}$
to the action space $\mathcal{A}$, so that the agent decides its action by choosing the optimal action for 
the current state regardless of its past history. This memory-less property is called \emph{Markov property} and is a fundamental assumption in such settings. Ultimately, 
the solution maximizes the expected total discounted reward:
\[ G_t  = E(\sum_{t=0}^{\infty}\gamma^{t}\mathcal{R}(s_t,\mathrm{a}_t,s_{t+1}))\]
It is important to note that $\gamma$ makes $G_t$ finite as it approaches zero given enough timesteps, which 
makes this definition appropriate even for tasks with infinite horizon.  

As the agent keeps interacting with the environment, 
it collects data in the form of trajectories of actions, 
observations and rewards. These experiences 
are utilized by the RL algorithm in order to update 
its current policy $\pi$ and make better decisions 
in the future. Under these circumstances, there 
are two main approaches to addressing 
decision-making challenges in the 
context of RL; 
\emph{model-based} methods and \emph{model-free} methods \cite{Nagabandi2017Neural,Xu2018Algorithmic}.

Model-based RL algorithms represent an important avenue of research. Such algorithms learn to simulate the world's dynamics, including its transition and reward functions. 
With this ability in hand, we can predict future transitions and utilize this knowledge to learn adequately, albeit with the burden of extra complexity. 
Therefore, model-based methods embed 
strategic foresight in the decision-making, and can attain greater 
efficiency in sample usage than model-free methods.  

Model-free 
methods, on the other hand, do not require a model of 
the environment to operate and rely 
solely on the experience gathered 
from the agent interacting with 
the environment. 
In turn, this family of algorithms  
can be classified into two frameworks; 
\emph{value-based} algorithms and 
\emph{policy-based} algorithms \cite{Arulkumaran2017Deep,Liu2020Overview}. 
In \emph{value-based} algorithms, the agent maintains a value 
function (usually in the form of a table) that maps 
state-action pairs ($s$, $\mathrm{a}$) (or just states) to a single, real, value. 
One of the most well-known algorithms in this category - 
and generally in RL - is $Q$-learning \cite{watkins1992q}. 
$Q$-learning, maintains a $Q$-table with values $Q(s,\mathrm{a})$. 
These $Q$-values serve as an estimate of the expected 
utility of taking action $\mathrm{a}$ in state $s$. This method dictates 
the agent to iteratively update its $Q$-values following the 
Bellman equation: 
\[Q(s,\mathrm{a}) \leftarrow Q(s,\mathrm{a})+\alpha \, (r+\gamma \, max_{\mathrm{a}'} Q(s', \mathrm{a}') - Q(s,\mathrm{a}))\]
where $\alpha \in (0,1]$ is the learning rate that controls how much we favor new information instead of old information. We can view this 
update as a convex combination of old and new information $Q(s,a) \leftarrow (1-\alpha) \, Q(s,\mathrm{a}) + \alpha \, (r+\gamma \, max_{\mathrm{a}'}Q(s',\mathrm{a}'))$ where $r+\gamma \, max_{\mathrm{a}'}Q(s',\mathrm{a}')$ is known as the \emph{target function}. 
Given enough updates of the $Q$-values, the agent can estimate accurately the utility of each available 
action at any given state. 
Nevertheless, a limitation of $Q$-learning is its inapplicability to environments with continuous representation.

On the other hand, \emph{policy-based} RL algorithms focus on maximizing directly the parameterized policy's rewards, 
through adjusting its parameters, instead of evaluating states. These methods often utilize an optimization algorithm to update the 
parameters efficiently, such as gradient ascent \cite{Brereton2009Steepest}, and therefore known more specifically as \emph{policy-gradient} algorithms. One classic 
algorithm of this family is REINFORCE \cite{Williams1992Simple}. REINFORCE 
updates the policy parameters based on 
the objective function $J(\theta)$: 
\[\nabla_\theta J(\theta) = E_{\tau \sim \pi_\theta} [\sum_{t=0}^T \nabla_\theta \,  log \, \pi_\theta (\mathrm{a}_t|s_t) \cdot G_t]\]
where $\tau$ is a sampled trajectory and $\pi_\theta$ is the current parameterized policy. $J(\theta)$ is used to update the parameters $\theta$ in the direction that maximizes the expected utility. This method has the advantage of being applicable in both 
discrete and continuous environments. 

\begin{figure}[h]
    \centering
    \includegraphics[scale = 0.6]{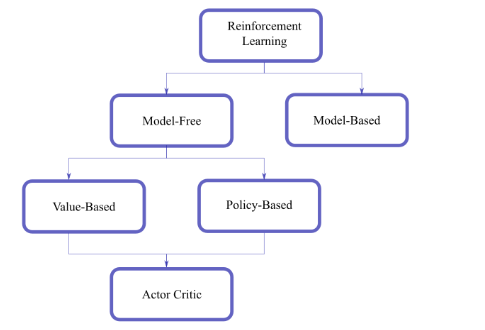}
    \caption{A taxonomy of RL algorithms \cite{canese2021multi}. Note that there exist many algorithms that are either policy-based or value-based and are not actor-critic.}
    \label{fig:rl_taxonomy}
\end{figure}

\emph{Actor-critic} RL algorithms combine the strengths of value-based and policy-based methods, offering a 
balanced approach for solving RL problems. In these algorithms, the ``actor'' updates the policy based 
on feedback from the ``critic'', which offers value estimation of the current policy. This synergy of different paradigms allows 
for more robust and efficient learning, and mitigates unilateral problems such as high variance and slow convergence. In Figure 
\ref{fig:rl_taxonomy} we can see a visualized taxonomy of the different families of RL algorithms, as presented in \cite{canese2021multi}.

Another way of categorizing RL algorithms is 
whether they are \emph{off-policy} or \emph{on-policy}. On-policy methods, like REINFORCE, use fresh data from the current policy in order to perform each update. 
In contrast, off-policy methods, like $Q$-learning,  can re-use old 
data generated by another policy, thus allowing 
for better sample-efficiency since these 
algorithms can utilize the same sample multiple times. The key for off-policy methods to work 
effectively, is to manage the deviations between 
the current policy and the old data-collecting 
policy.

A very important notion in RL, is the 
exploration-exploitation dilemma. Since in this 
context we aim to compute an optimal policy, it is 
crucial for any RL algorithm to be able to keep 
learning better policies and not get stuck in local optima. In other words, the algorithm must choose when to exploit the current 
best action or explore with the hope of finding 
a better one. A popular, yet simple, strategy that balances exploration 
and exploitation is the $\epsilon$-greedy. This strategy utilizes 
randomness by choosing to explore a random action with a 
probability $\epsilon$ or to exploit the best known action $\mathrm{a}^*$, 
according to the model's current estimates, with probability 
$(1-\epsilon)$: 

\[\mathrm{a}_t = \begin{cases}\mathrm{a}_t^* & \text{, with probability } 1-\epsilon_t \\ \mathrm{a} \sim \mathcal{U} & \text{, with probability } \epsilon_t \end{cases}\]
Typically, we aim for our agent to explore more in 
the early stages of training and exploit more as training progresses. So it seems only natural to anneal the $\epsilon$ 
probability from a high initial value (usually 1) to a low 
value (usually $\leq 0.1$) as training progresses. 
More sophisticated exploration strategies exist with potentially better performance than $\epsilon$-greedy, such as 
\emph{Boltzmann} (also referred to as \emph{Softmax}) Exploration \cite{cesa2017boltzmann}, \emph{Upper Confidence Bound} \cite{auer2002finite} and \emph{Thompson Sampling} \cite{Russo2017A}.

Despite their efficiency and rigorousness, RL algorithms suffer from the \emph{curse of dimensionality}. 
As the potential states and actions of the environment increase, RL algorithms have a hard time 
exploring and learning effective policies. The potential vastness of the problem's dimensions can make the agent encounter novel states more frequently and have fewer opportunities to learn 
from repeated situations in order to capitalize on any good policies found through exploration. 

This serious 
challenge can be partially overcome with the use of non-linear function approximation \cite{Goddard1965Approximation} NN-based techniques, such as {\em deep learnign (DL)} methods~\cite{Deng2014Deep,LeCun2015Deep}. DL involves deep NNs with many hidden 
layers, typically ranging from at least one to often two to ten hidden layers. This structure makes the model able to adapt to more complex functions and capture high-level features in input data. As expected, 
in recent years, methods using DL have excelled in intricate problems with high dimensionality such as 
image recognition and natural language processing, including advanced applications like large language models \cite{Zhao2023A} and generative adversarial networks \cite{Goodfellow2014Generative} that produce human-like text and highly realistic images. 

To this end, DL and NNs have given rise to deep reinforcement learning (DRL),  enabling RL to tackle 
problems that were previously deemed infeasible due to the curse of dimensionality.  Of course, most of the 
aforementioned 
core frameworks 
of classic RL, such as MDPs, 
POMDPs, algorithm categories, 
solution concepts and exploration 
strategies, transfer over to the domain 
of  DRL. 

In recent years, research in DRL has sparked off a wide array of  methodologies that push the boundaries of the field. By combining DL with 
already-existing RL literature several advancements have been made.  These advancements include policy-based algorithms 
like Proximal Policy Optimization (PPO) \cite{schulman2017proximal} and value-based 
algorithms like Deep $Q$-Network (DQN) \cite{Mnih2013Playing,Mnih2015Human-level}. Also,  
actor-critic models that intertwine value and policy based methods to benefit from both, have been formulated, such as Soft Actor-Critic (SAC) \cite{haarnoja2018soft} and Deep Deterministic Policy Gradient (DDPG) \cite{lillicrap2019continuous}. 
These new approaches have also led to more 
advanced techniques such as \emph{imitation learning}, where 
agents can observe expert demonstrations to learn faster. Algorithms from the field of DRL have since been applied to  
a diverse plethora of problems, both discrete and continuous, 
such as robotics \cite{Gu2016Deep,kober2013reinforcement}, Atari games \cite{Mnih2013Playing,mnih2015human}, the board game of Go 
\cite{Silver2016Mastering}, and autonomous driving \cite{karalakou2023deep,Sallab2017Deep,Wang2018Deep}, to name a few.

We now briefly present a few popular and diverse 
DRL algorithms that have contributed vastly in the field; DQN, 
PPO, SAC.
Specifically, DQN is a model-free, off-policy, value-based 
algorithm that maintains an NN to 
approximate the $Q$-value function given a state-action 
pair. It stores encountered state transitions and reward 
outcomes in an experience replay buffer that is then used 
to sample from and to do gradient updates on the NN. 
For each update we calculate the loss between the current 
estimates and the target function and backpropagate it through 
the network. More often than not, an $\epsilon$-greedy strategy 
is used during action selection. 

A lot of algorithms 
have stemmed from DQN in recent years with remarkable 
results in discrete domains. 
As an example, \cite{orfanoudakis2023novel} 
puts forward a novel approach to managing distributed 
energy resources in the context of the \emph{Smart Grid}
 by employing DQN to predict the most beneficial 
combinations of local flexibility estimators. Another notable work is that of \cite{chrysomallis2023deep}, which enhances DQN with {\em implicit imitation} learning to give rise to the so-called {\em Deep Implicit Imitation Q-Networks (DIIQN)} algorithm.
The approach makes use of {\em only} state transitions, obtained from a mentor, to accelerate learning, without requiring access to observations of the mentor actions; 
and can be applied in a host of domains, including autonomous driving. That work was later extended to allow for imitation learning when the mentor and observer agent action sets are {\em heterogeneous}, for the first time in the literature~\cite{chrysomallisAAAI2025heterogeneous}.

An important spin-off of DQN is  
Bootstrapped DQN \cite{osband2018randomized,osband2016deep} that employs several network heads to benefit from the 
diverse outputs they generate. Essentially, instead of learning 
a single policy, Bootstrapped DQN learns multiple 
policies by utilizing the inherent uncertainty of the distinct heads. 
At the start of each episode, a head is sampled  
uniformly at random, to act on the environment 
and collect data. The data collected can be stored in the same replay buffer, but a 
random distribution is used to dictate which heads 
are to use each experience sample.  
This method has been shown to 
outperform $\epsilon$-greedy, especially in problems 
that require deep exploration.

Another popular algorithm in the related literature is PPO. PPO operates as a model-free, on-policy, policy-based algorithm that 
employs stochastic gradient ascent to optimize the policy directly, instead of using $Q$-value 
estimates like DQN. 
This method aims to maximize a clipped surrogate objective 
function which leads to the avoidance of 
drastic changes and maintains a more stable learning pace. This simple and straightforward 
technique balances well the trade-off between exploration and exploitation. The robustness 
and stability of PPO enables it to attain high performance in a variety of domains 
(both discrete and continuous), while not requiring any extended search for hyperparameter values. 

Last but not least, SAC is a model-free, off-policy, actor-critic algorithm that makes use of the entropy in 
the action selection procedure, giving rise to a more efficient exploration method compared to 
$\epsilon$-greedy. It can be naturally applied to continuous environments and produces a stochastic 
policy by design. SAC also maintains a 
replay buffer as an off-policy algorithm, and two networks based on an actor-critic architecture. The key innovation of this algorithm is the augmentation of an entropy term in 
the standard intrinsic rewards from the environment, that naturally balances exploration and 
exploitation. This makes SAC more sample efficient than 
most traditional DRL algorithms.

Despite the achievements in the field of DRL, there are still several problems that 
limit its use in real-world problems. Some of the most important issues are \emph{sample inefficiency}, 
\emph{inability to generalize} to new states/problems, the demand for \emph{extensive computational 
resources} and the \emph{obligation for a simulator} to exist. Sample inefficiency refers to the requirement 
of vast amounts of data to even learn a substantially good policy. Even algorithms that are deemed 
sample-efficient are not efficient enough to be applied to many real-world tasks that cannot offer numerous  
samples or that even acquiring new samples is expensive. Additionally, most algorithms cannot generalize 
their knowledge to other domains or even to underexplored parts of the same environment. 
This lack of generalization means that DRL methods need to be trained again even if the environment is slightly different - a burden that can be resource exhaustive. As it happens, training a DRL algorithm, often with more than one neural networks, demands a large amount of computational resources 
for extended periods of time, often several days. Consequently, this issue only, can make DRL 
impractical for individuals that do not have access or cannot afford such computational power.  
Finally, many physical problems can be proven to be expensive to train an agent in, especially 
tasks in which the agent has a physical substance. For example, if the task is to train a real-world physical 
robot, the decisions made by the algorithm can damage the agent. For these tasks there need to be a simulator that can simulate both the agent and the environment in a digital form to allow for a faster 
and safer training procedure. However, as is evident, such simulators are not always available and their  
implementation introduces an additional hurdle. 

\subsection{Multiagent Reinforcement learning}

\begin{figure}[hb!]
    \centering
    \includegraphics[scale = 0.8]{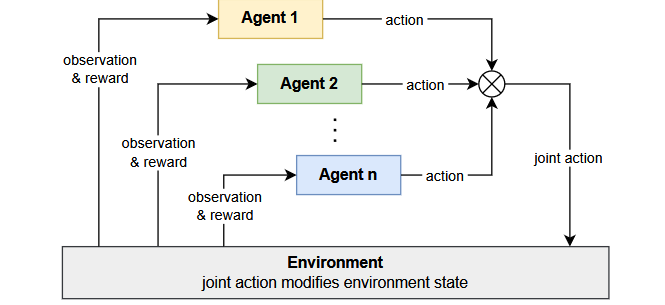}
    \caption{The MARL setting~\cite{marl-book}.}
    \label{fig:marl_scheme}
\end{figure}

A multiagent learning direction that has gained momentum over the last years 
is {\em multiagent reinforcement learning (MARL)}~\cite{marl-book,gronauer2022multi,oroojlooy2023review}. MARL is an extension of 
single-agent RL and it 
involves multiple agents that usually interact with each other 
and affect each other's decisions, as well as each other's 
perception of the environment. The relationships between the 
agents may vary as they may need to cooperate, compete or adopt a 
mix of both behaviors, depending on the task at hand. As expected, these new dynamics that do not 
exist in single-agent settings, mandate the incorporation of 
game theory notions and techniques in the RL setting to model these more complex MARL notions and  enable the analysis of strategies and outcomes. Naturally, just as DL 
extends the capabilities of RL, 
it has similarly enhanced MARL 
to design {\em multiagent} DRL (MADRL) algorithms 
that are more effective in solving 
real-world problems. 
Typically, many 
real-world scenarios can normally be modeled with the use of multiple agents 
such as in autonomous driving \cite{shalev2016safe}, other settings with a multitude of economic 
entities \cite{shavandi2022multi}, or smart-grid control \cite{Lu2020Multi-agent}, to name a few. 
All these scenarios benefit from the ability to 
model multiple interconnected agents and their 
strategic interactions. 
A schematic overview 
of the MARL setting is depicted in 
Figure~\ref{fig:marl_scheme}~\cite{marl-book}.

{\em Markov games (MGs)} \cite{marl-book,littman1994markov,littman2001value,shapley1953stochastic} 
(also known as \emph{stochastic games}), have been key for the formal definition of the MARL framework. They can be considered as 
an extension of MDPs to contain multiple decision-making entities in a common 
environment. Specifically, MGs allow for multiple co-existing agents that take 
actions simultaneously, and encapsulate various interactions among them---cooperative, competitive or even a mix of both. 
Formally, an MG is a ($\mathcal{I}$, $\mathcal{S}$, $\{\mathcal{A}^i\}$, $\mathcal{T}$, $\{\mathcal{R}^i\}$, $\gamma$, $\mu$) tuple~\cite{marl-book}, where: 
\begin{itemize}
    \item $\mathcal{I}$ denotes the set of agents that coexist in the environment
    \item $\mathcal{S}$ denotes the state space of the environment
    \item $\mathcal{A}^i$ is the action space for agent $i$ and $\mathcal{A}=\mathcal{A}^1\times\mathcal{A}^2\times \ldots \times\mathcal{A}^n$ represents the joint action space. A set of selected actions can then be denoted as $\mathrm{a}$ 
    \item $\mathcal{T}(s, \mathrm{a}, s')$ = $\Pr(s'|s,\mathrm{a})$ is the transition function that maps state and joint action ($s$,$\mathrm{a}$) pairs to a probability distribution for each possible next state $s'$
    \item $\mathrm{R}^i(s,\mathrm{a},s')$ is the reward function for agent $i$, based on the current state $s$, the action $\mathrm{a}$, and the next state $s'$
    \item $\gamma \in [0,1]$ is the discount factor
    \item $\mu: S \xrightarrow{}[0,1]$ is the initial state distribution
\end{itemize}
Despite the importance of MGs in the field, this framework does not account for 
partial observability in the environment from the perspective of each agent. This gap 
is bridged by an extension of MGs, the {\em Partially Observable Markov Games (POMGs)}~\cite{marl-book}. 
In the context of POMGs, each agent operates in a subjective way, based on its 
individual view of the world. Considering that its point of view is likely 
to be incomplete due to other agents being part of the environment, the importance of  
frameworks that account for partial observation is apparent.  

Formally, a POMG can be represented 
as a tuple ($\mathcal{I}$, $\mathcal{S}$, $\{\mathcal{A}^i\}$, $\mathcal{T}$, $\{\mathcal{R}^i\}$, $\{\Omega^i\}$, $\{\mathcal{O}^i\}$, $\gamma$, $\mu$)~\cite{marl-book} where: 
\begin{itemize}
    \item $\mathcal{I}$, $\mathcal{S}$, $\mathcal{A}^i$, $\mathcal{T}$, $\mathcal{R}^i$, $\gamma$, $\mu$ are the same as in MGs
    \item $\Omega^i$ denotes the observation space for agent $i$ 
    \item $\mathcal{O}^i(s',\mathrm{a}, \omega^i)=\Pr(\omega^i|s',\mathrm{a})$ where $\omega^i \in \Omega^i$, is the observation function; an agent $i$ receives an observation according to $\mathcal{O}^i$ after making a transition in the environment
\end{itemize}
In multiagent settings, at each timestep $t$, all $n$ agents choose an action $\mathrm{a}^i_t$ and perform all the actions simultaneously. After taking that joint action $\mathrm{a}_t$ each one receives an observation $\omega^i_{t+1}$ with probability $\mathcal{O}^i(s_{t+1},a_t, \omega^i_{t+1})$ and an individual reward $\mathrm{r}^i_t$ from $\mathcal{R}^i(s_t,\mathrm{a}_t,s_{t+1})$ while the environment transitions from current state $s_t$ to the next state $s_{t+1}$ with a probability corresponding to $\mathcal{T}(s_t,\mathrm{a}_t, s_{t+1})$. At $t=0$ specifically, 
an initial state $s_0$ is sampled 
from $\mu$.

Noticeably, the differences between POMGs and MGs are analogous to those between MDPs and POMDPs, albeit the latter do not account for multiple  agents in the environment. 
To this end, {\em Decentralized POMDPs (Dec-POMDPs)} ($\mathcal{I}$, $\mathcal{S}$, $\{\mathcal{A}^i\}$, $\mathcal{T}$, $\mathcal{R}$, $\{\Omega^i\}$, $\{\mathcal{O}^i\}$, $\gamma$, $\mu$)
constitute a special case of POMGs, with the sole 
difference that agents share a common reward function $\mathcal{R}$~\cite{marl-book}: 
\begin{itemize}
    \item $\mathcal{I}$, $\mathcal{S}$, $\mathcal{A}^i$, $\mathcal{T}$, $\Omega^i$, $\mathcal{O}^i$, $\gamma$, $\mu$ are the same as in POMGs
    \item $\mathcal{R}$ is the 
    global reward function, common to all agents 
\end{itemize}
The absence of distinct individual reward functions dictates a 
cooperative aspect, as all agents seek to maximize the 
global rewards. 
The solution to a Dec-POMDP is considered as a joint policy $\mathbf{\pi}^*$ that maximizes the expected collective sum of discounted rewards $\max_{\mathbf{\pi}} E(\sum_{t=0}^\infty \gamma^t R(s_t,\mathrm{a}_t,s_{t+1}))$. Note that $R(s_t,\mathrm{a}_t,s_{t+1})$ here is the global reward received by all agents after the transition from timestep $t$ to timestep $t+1$. 

With the deployment of multiple agents in RL, several new challenges emerge. An evident issue is 
that of non-stationarity that results from the 
introduction of multiple decision-making entities 
in the environment as each agent's actions change 
the context for other agents. This results in a more erratic learning process as agents maintain partial observability over the environment. Another challenge is that it 
is often difficult to evaluate the contribution of 
each agent to the outcome and, thus, to reward each 
agent fairly. This is widely known in the literature 
as the \emph{credit assignment problem}. Furthermore, the 
addition of numerous agents within the RL setting
results in an exponential growth in complexity,  
limiting this way the scalability of most MARL 
algorithms. 

Since we now maintain numerous decision-makers, 
it is crucial to decide how the decisions are 
being made from a technical perspective. There 
are two notable approaches, the centralized 
decision and the distributed decision making. 
In the context of the centralized decision-making, 
a central entity is tasked with producing joint 
actions $\bm{\mathrm{a}}$ based on the state of the 
environment. On the other hand, each agent can 
compute its own action based on its local 
observation of the state. The latter, normally 
enhances scalability, but often requires the use 
of sophisticated coordination mechanisms. As a consequence, 
the design of such mechanisms seems necessary, 
especially for cooperative scenarios where the 
agents are expected to reach a common goal. 
Another critical notion in MARL is that of agent modeling. Agent modeling 
focuses on designing, maintaining and updating models that are used to predict 
the future actions of other agents' in the environment.  This can lead to a 
more successful learning process since agents can anticipate the actions of 
other agents and adjust their strategies preemptively. The importance of agent 
modeling is essential in competitive scenarios (in that case it is also known 
as opponent modeling) where being able to forecast opponents' moves can 
provide a strategic advantage. 

After establishing the framework of operation for learning in MAS and discussing potential problems 
as well as techniques employed, we now examine a few selected algorithms 
in detail.

In a comprehensive analysis, \cite{jiang2018learning}  
notes that communication in cooperative tasks is essential, 
but constant communication may impose problems in performance and energy efficiency. Consequently, they propose a model 
named ATOC, that aims to learn when communication is 
indeed necessary and how to pass information effectively. 
ATOC, in essence, dictates the agents to form dynamic 
communication groups to enhance their coordination. 
In more detail, this algorithm trains an \emph{attention unit} with 
local observations and action intentions of agents to 
learn when communication is imperative. When the 
attention unit decides that communication is due, a bi-directional Long Short-Term Memory (LSTM) unit serves as a communication channel. 
This attention-based communication mechanism is 
integrated in an actor-critic model that is trained in a 
centralized manner, although each agent makes individual observations. In addition, this algorithm does not assume the SIH, 
since agents may be required to maximize the collective utility, depending on the 
problem at hand. 
The evaluation of ATOC across various cooperative 
problems, shows that it outperforms existing methods by 
enabling agents to coordinate in a more sophisticated 
manner.

Another notable paper is \cite{chu2019multi} that introduces a fully scalable and decentralized MADRL 
algorithm, the Multiagent Advantage Actor Critic (MA2C). MA2C is built as an extension to the well-known DRL algorithm Advantage Actor Critic (A2C) and is designed to tackle the adaptive traffic signal control 
problem. It does so, by enabling limited communication between the agents, and specifically
sharing fingerprints of policies between neighbors. The connections between the agents help to alleviate the issues caused by partial observability and non-stationarity, as it helps each entity to 
have more informed decision-making based on the anticipated actions of its neighbors. Another key 
feature of this framework, is that it tailors the reward of each agent $i$, by computing a discounting, weighted, global reward 
with increased weights for the rewards of agents in agent's $i$ neighborhood. Essentially, each individual's 
reward is discounted according to the agent's distance from agent $i$. 
MA2C is evaluated on both a large synthetic traffic grid and the real-world traffic network of Monaco city. 
In this context, the algorithm demonstrates superior performance of optimality, robustness and sample 
efficiency over traditional decentralized MADRL algorithms and other state-of-the-art methods.

Despite the fact that they can be criticized as lacking in theoretical foundations in a multiagent setting, independent learners (IL) can attain high performance 
in hard multiagent problems. The underpinning of 
IL algorithms is that each agent learns on its own and perceives the rest of the agents as part of the 
environment with no communication in between, thus pertaining to the SIH. This simple logic can surprisingly attain performance comparable, and sometimes even 
better than more sophisticated techniques \cite{de2020independent,papoudakis2020benchmarking,yu2022surprising}, and simultaneously be scalable due to its by-design decentralization. The first algorithm introduced in this 
context is a value-based algorithm based on $Q$-learning, the Independent $Q$-learners (IQL) \cite{tan1993multi}. According to IQL, each agent maximizes
their own $Q$-function independently and optimize 
their own policy without sharing any 
information and without cooperating. 
A more modern approach adapts PPO in independent multiagent environments to present the 
actor-critic algorithm called Independent Proximal Policy Optimization (IPPO) \cite{de2020independent}.  
In a similar fashion to IQL, IPPO does not require any information sharing or communication; instead, each 
agent works towards optimizing its own policy. Despite that, IPPO has achieved 
very good performance in hard problems such as the Starcraft Multiagent Challenge. 
Although IL can achieve 
competitive results and at the same time be 
straightforward and simplistic, it may struggle 
with the non-stationarity of the environment 
that is introduced by other agents' behavior, since each decision-maker acknowledges its 
equivalents as dynamic parts of the environment.

A relevant algorithm that uses a more sophisticated approach and is based on PPO, 
is that of Multiagent Proximal Policy Optimization (MAPPO) \cite{yu2022surprising} that achieves good results in 
several cooperative benchmarks. MAPPO is an actor-critic algorithm, that follows 
a centralized training architecture which in turn 
leverages global information as the critic is trained on the joint state value function, making the agents more altruistic and disregarding the SIH. Additionally, there is no communication incorporated in 
this technique since the agents train in a centralized manner. 
The result is a 
sample-efficient algorithm that, unlike IL 
approaches, mitigates the non-stationarity 
problem making it less susceptible to 
changes in the environment made by other 
learners. As its single-agent counterpart, 
MAPPO serves as a very strong baseline 
due to its high performance in most cooperative 
environments and its robustness to 
hyperparameter tuning \cite{papoudakis2020benchmarking}. 

Another interesting approach is the Value Decomposition Networks (VDN) 
\cite{sunehag2017value}. VDN is a groundbreaking algorithm that is designed to 
train multiple agents to work together towards 
achieving a common goal. Its key point, as the name suggests, is the principle 
of decomposing the joint value function into a 
sum of individual agents' value functions. 
Ultimately, each agent builds a policy that 
maximizes the sum of policies which serves as 
a proxy for the joint value function. 
This 
offers a subtle way of coordination without the 
use of explicit communication mechanisms, 
since each entity makes decisions based on its 
perspective while indirectly contributing to 
collective intelligence. By utilizing the sum of these individual contributions and matching local rewards to the corresponding agents, VDN tackles two 
inherent MAS pathogenies such 
as the credit assignment problem and scalability. 
This method has opened up new avenues in 
research for designing sophisticated MAS capable of complex interactions and 
deeply strategic behavior. 

Finally, the Cooperative Multi-Agent Exploration (CMAE) \cite{liu2021cooperative} 
is an exploration strategy that works on top of any MARL algorithm and 
promotes coordination through shared 
goals, particularly in sparse-reward environments. CMAE innovates by projecting the complex, high-dimensional state space into simpler, lower-dimensional ``restricted spaces'' for more efficient exploration.  
Using a hierarchical approach, exploration starts from these simpler spaces and expands as needed. 
This mechanism reshapes the rewards 
within the replay buffer to incentivize agents towards 
common objectives. 
It relies on entropy-based 
techniques for goal selection and demonstrates significant 
improvements in learning efficiency and effectiveness in tasks within environments like the Multiple 
Particle Environment and the StarCraft Multi-Agent Challenge, showcasing its superiority over 
traditional exploration methods. 

In MAS, the coexistence of multiple agents renders 
the environment strategic, in the sense that it can be modeled 
as a stochastic game, since the decisions of other agents affect the environments predictability and bear (long-term) consequences . Some of the algorithms reviewed 
in this section are essentially popular single-agent DRL 
algorithms, such as MA2C and MAPPO, adapted in MA 
scenarios. Other algorithms, such as IQL and IPPO, are simply 
multiple instances of single-agent methods running in 
parallel where agents interact within a common environment 
yet do not acknowledge each other's existence. 

In the context 
of MARL techniques that depend on deep NNs, the 
reliance in CPA is a subtle discussion, in the sense that 
they cannot assimilate any amount of knowledge in the 
form of common priors. 
Usually, we can assume that the network 
weights are initialized arbitrarily so that each agent may incorporate 
any beliefs of its own, while disregarding other agents' beliefs. 
Traditionally, the CPA underpins theoretical 
frameworks and models, suggesting that 
agents share a base of common knowledge 
that serves as a universal starting point. 
The CPA, albeit convenient in analysis and 
algorithm design, does not always reflect the 
intricacies and complexity of many real-world 
scenarios, where it may seem natural for agents 
to have distinct beliefs. 

As noted early on in this document, another widely used assumption 
is the SIH which proclaims that 
each agent's objective is the maximization of its own 
potential payoff. 
Identifying this gap, 
we aim to incorporate a phase in the 
learning process for each agent to formulate 
its own belief based on the context of the 
environment containing the other agents. This 
approach does not only add a layer of realism to 
multi-agent settings, but also enriches the 
strategic aspect of such games and amplifies the 
need for connections between the agents. 
We argue that the formulation of personalized and potentially 
heterogeneous beliefs could lead to varying results in 
different benchmarks and deem the discarding of the 
CPA and the SIH pivotal in multi-agent domains. 
The development of equilibria and rationality solution concepts 
that echo their independence from the SIH and the CPA, seems imperative. 

In Table \ref{tab:madrl_taxonomy}, we sum up the MARL algorithms discussed in this section, 
by underscoring common features as well as key 
differences.

\begin{table}[h]
\centering
\resizebox{0.95\textwidth}{!}{%
\begin{tabular}{c|cccccc|}
\cline{2-7}
\textbf{} &
  \multicolumn{6}{c|}{Algorithms} \\ \cline{2-7} 
 &
  \multicolumn{1}{c|}{\textbf{MA2C} \cite{chu2019multi}} &
  \multicolumn{1}{c|}{\textbf{ATOC} \cite{jiang2018learning}} &
  \multicolumn{1}{c|}{\textbf{VDN} \cite{sunehag2017value}} &
  \multicolumn{1}{c|}{\textbf{MAPPO} \cite{yu2022surprising}} &
  \multicolumn{1}{c|}{\textbf{IQL} \cite{tan1993multi}} &
  \textbf{IPPO} \cite{de2020independent} \\ \hline
\multicolumn{1}{|c|}{\textbf{Decentralized Training}} &
  \multicolumn{1}{c|}{\ding{51}} &
  \multicolumn{1}{c|}{\ding{55}} &
  \multicolumn{1}{c|}{\ding{55}} &
  \multicolumn{1}{c|}{\ding{55}} &
  \multicolumn{1}{c|}{\ding{51}} &
  \ding{51} \\ \hline
\multicolumn{1}{|c|}{\textbf{Communication during the learning task}} &
  \multicolumn{1}{c|}{\ding{51}} &
  \multicolumn{1}{c|}{\ding{51}} &
  \multicolumn{1}{c|}{\ding{55}} &
  \multicolumn{1}{c|}{\ding{55}} &
  \multicolumn{1}{c|}{\ding{55}} &
  \ding{55} \\ \hline
\multicolumn{1}{|c|}{\textbf{Policy-gradient Method}} &
  \multicolumn{1}{c|}{\ding{51}} &
  \multicolumn{1}{c|}{\ding{51}} &
  \multicolumn{1}{c|}{\ding{55}} &
  \multicolumn{1}{c|}{\ding{51}} &
  \multicolumn{1}{c|}{\ding{55}} &
  \ding{51} \\ \hline
\multicolumn{1}{|c|}{\textbf{Value-based Method}} &
  \multicolumn{1}{c|}{\ding{51}} &
  \multicolumn{1}{c|}{\ding{51}} &
  \multicolumn{1}{c|}{\ding{51}} &
  \multicolumn{1}{c|}{\ding{51}} &
  \multicolumn{1}{c|}{\ding{51}} &
  \ding{51} \\ \hline
\multicolumn{1}{|c|}{\textbf{Does not rely on SIH}} &
  \multicolumn{1}{c|}{\ding{51}} &
  \multicolumn{1}{c|}{\ding{51}} &
  \multicolumn{1}{c|}{\ding{51}} &
  \multicolumn{1}{c|}{\ding{51}} &
  \multicolumn{1}{c|}{\ding{55}} &
  \ding{55} \\ \hline
\end{tabular}%
}
\caption{A taxonomy of the MARL algorithms covered, in the context of their features}
\label{tab:madrl_taxonomy}
\end{table}

\section{Rationality Solution Concepts}
\label{sec:equil}
In complex multi-agent domains, understanding Nash
equilibrium (NE) and other  rationality solution concepts is vital 
for designing agents that interact effectively, 
as well as analyzing and providing insights on 
their behavior and outcomes. 
Usually, the desired properties of the outcomes in this scope are fairness and stability. Fairness corresponds to how well the agents' payoffs comply with their actual contributions, while stability reflects the non-willingness of agents to select a deviating strategy~\cite{chalkiadakis2022computational}.
In this section, we will present and discuss 
such game theoretic notions within the field 
of multi-agent DRL, with an emphasis on correlated equilibrium 
(CE), and on fairness-related solution concepts. 

To begin with, the well-known Nash theorem guarantees that every finite two-player game has {\em at least} one NE. In games where multiple equilibria exist, one has to search for the most desirable one among them.
One measure of desirability is the degree of social welfare maximization, i.e. the sum of all players' payoffs.
The search for an NE is PPAD-complete, however finding an NE with maximal social welfare is NP-hard even in symmetric 2-player games, making it computationally challenging. There have been attempts to reach approximate Nash equilibria that achieve near optimal social welfare. In~\cite{czumaj2015approximate} it is shown that this can be discovered in polynomial time. 

Now, CE offer a perspective often broader than 
vanilla NE, by introducing the 
coordination between agents through external 
signals, thus making the agents' strategies 
correlated, in contrast with NE where each 
agent's strategy is independent of the others. 
CE allow for more efficient coordination among 
agents, potentially leading to better outcomes. 
In the specific context of multi-agent DRL, utilization of CE 
can be crucial especially in complex environments 
that require a high level of coordination and 
encapsulate intricate interconnections. 

Moreover, there exist certain successful integrations 
of CE in multi-agent DRL. As an example, 
in \cite{yu2014multi} the authors apply CE on smart generation 
control in power systems. They present a novel 
algorithm that learns to achieve optimal 
joint equilibrium strategies. This approach 
employs the correlated equilibria in order to 
facilitate optimal coordination and consequently 
improve performance. Similarly, in \cite{gan2019new} they 
employ a dynamic correlation matrix to model 
agent interactions and complex relationships. 
With this technique, the authors enable the 
discovery of optimal CE and enhance collective 
outcomes. That work highlights how valuable are 
CE in hostile environments in which agent 
interconnections are complex and change dynamically. Another successful application 
is presented in \cite{he2022multi}, where  
multiple agents employ the DQN 
algorithm to achieve CE optimal solutions 
in order to optimize a textile 
manufacturing optimization process. 

In~\cite{graf2024computational}, authors employ DRL for training agents that participate into uniform price auctions. Interestingly, this Deep Deterministic Policy Gradient (DDPG) 
algorithm leads to learned policies which are shown to converge to NE. 
Also, this algorithm can tackle continuous state and action spaces, a feature that Q-learning does not possess, though a drawback of this approach is that it requires fine tuning of the related parameters.

We now turn our focus to the analysis of solution concepts produced by multiagent DRL 
algorithms, considering their fairness under the lens of game theory. 

First of all, fairness is crucial for ensuring that multiagent systems 
are efficient and stable, and can be encapsulated 
in game theoretic concepts. 
The importance of fairness is underscored in 
real world problems such as traffic control and 
network routing as they demand and equitable 
allocation of resources among the entities in 
the environment. Meanwhile, environment 
aspects such as partial information and dynamic features pose a challenge to evaluating 
fairness in multiagent DRL algorithms.  

A work that addresses fair solutions is \cite{jiang2018learning}. 
In that work, authors present a novel 
Fair-Efficient Network (FEN) model that 
manages to learn both efficiency and 
fairness in multiagent domains. This is 
achieved via fairness decomposition 
into objectives and utilizes a hierarchical structure, which contains 
a controller and multiple sub-entities, 
each one maintaining a distinct policy. 
This method supports fairness and 
addresses the challenges of stability in 
learning environments, as it can scale 
to larger problems due to the model's 
ability to facilitate decentralized learning. Another notable contribution, 
is that of \cite{zimmer2021learning}, where 
a metric is put forward that 
evaluates fairness in the form of 
optimizable welfare functions. A novel NN architecture is used
that consists of sub-networks and is 
capable of optimizing the different 
aspects of fairness such as efficiency and 
equity in the form of welfare functions. 
Ultimately, this technique leads to fair 
outcomes. Additionally, in \cite{wang2020multi}, a novel framework 
for controlling Unmanned Aerial Vehicles (UAVs) and optimizing their 
trajectories in terms of both performance and fairness is presented. In this context, 
authors propose a method where each UAV maximizes its reward independently and, at the same time, a collective approach for maximizing fairness is maintained.

Solution concepts play a pivotal role in generating 
strategies that are fair and robust. 
It is evident that CE and fairness can play an important role in the analysis and advancement of 
multiagent DRL. CE offer a promising avenue for addressing the coordination challenges of 
complex and, potentially, dynamic environments, and on the other hand, maximizing fairness of 
the solution concepts can amplify the quality of the generated strategies. Crucially, the application 
of these concepts to solve real-world problems further underscores their significance. 

Finally, we list the most common solution concepts that can be found in the MARL literature. These are the Shapley value~\cite{shapley1953value}, the Banzhaf index~\cite{banzhaf1964weighted}, the core~\cite{gillies1959solutions}, the nucleolus~\cite{schmeidler1969nucleolus}, the kernel~\cite{davis1965kernel}, the bargaining set~\cite{aumann1964bargaining}, and the stable set~\cite{von1947theory}.

The Shapley value guarantees that an agent will receive a payoff that is proportional to its contribution. Considering a characteristic function game $G=(N,v)$ with $N$ agents and $v$ the characteristic function, then the Shapley value $\phi_i(G)$ is calculated based on each agent's $i$ marginal contribution:
$$\phi_i(G) = \frac{1}{|N|!}\sum_{\pi \in \Pi_N} v(S_{\pi}(i)\cup\{i\})-v(S_{\pi})$$
where $\pi \in \Pi_N$ is a permutation of the agents among all permutations $\Pi_N$, and $S_{\pi}(i)$ the set of all predecessors of agent $i$ in $\pi$.

The Banzhaf index $\beta_i(G)$ is similar to the Shapley value, but instead of considering the average of all the possible agent permutations, it does so over all coalitions in the game:
$$\beta_i(G) = \frac{1}{2^{n-1}}\sum_{C \subseteq N \backslash\{i\}}[v(C\cup\{i\})-v(C)]$$

While the Shapley value and the Banzhaf index focus on the fairness of the payoffs, there are also solution concepts that deal with the stability of the formed coalitions in the game.
The core solution concept is the set of all outcomes $(CS, \mathbf{x})$, where $CS$ is a coalition structure and $\mathbf{x}$ the payoff vector, for which $x(C)\geq v(C)$ for every $C\subseteq N$, where $x(C)$ denotes the total payoff for $C$ under $\mathbf{x}$. Note that not every game has a non-empty core.

On a `reverse' direction, the nucleolus focuses on the payoff deficit $d(\mathbf{x},C) = v(C) -x(C)$ that a coalition would have in an outcome $x(C)$ with respect to its optimal payoff $v(C)$.
The nucleolus is guaranteed to be non-empty for any superadditive game, and in particular it includes exactly one element. 
Now, the kernel, which includes all outcomes for which no player can credibly demand a fraction of another player's payoff, includes the nucleolus, and is thus non-empty.
Finally, the bargaining and stable sets concern all the payoff vectors for which no agent would justifiably object to, and the extension to a set-valued solution concept, respectively.

The works of~\cite{chalkiadakis2007bayesian,10.1145/1329125.1329203} combined the concept of Bayesian core, a cooperative solution concept, with non-cooperative solution concepts. 
This is inline with the our objectives, since we envision to revolutionize the way we think about equilibria, and pave the
road towards the discovery of novel, meaningful, and practical concepts of rationality, so as to enable
computationally tractable strategic deliberations in uncertain multiagent environments. 
Recall that a grand challenge put forward early in this paper, is that of  
overriding restricting GT assumptions, such as the SIH and CPA,\footnote{Note that the Bayesian core solution concept does not necessitate the use of the CPA assumption.} and to identify tractable ways for
computing initial heterogeneous agent beliefs into the focus. 
This could potentially give rise to novel, practical equilibrium concepts, which explicitly take the initial
agent beliefs-forming stage into account - and which come complete with bounded rationality capabilities
and anytime performance guarantees. 
Such equilibrium concepts would have
the potential to be employed in cooperative and non-cooperative domains alike, rendering as such the
distinction between cooperative and non-cooperative GT solution concepts artificial and obsolete.

\section{Probabilistic Topic Modeling}
\label{sec:PTM}

Existing Probabilistic Topic Modeling (PTM) algorithms have been applied successfully in various domains, such as document and text classification, image analysis and organization, movie recommendation systems, and population genetics~\cite{blei2012}.
One could potentially adapt existing PTM algorithms for opponent modeling or strategy modeling purposes in multiagent environments---i.e., use them to derive the prior beliefs of the agents. 
Note that the ability to assign meaningful probability to previously unseen ``documents'' is inherent in many existing PTM algorithms (such as the LDA). Note also that PTMs constitute Bayesian exploration techniques, assigning as they do ``topics’’ to modeled items according to probabilistic beliefs embodied in the mixture distributions they maintain. As such, they do not require the use of explicit heuristics or stoppage conditions to tackle the exploration-exploitation trade-off that is ever-present in decision-making problems.

Probabilistic topic models~(PTMs)
 utilize statistical methods to analyze words in
documents, so as to discover the underlying topics and their interconnection.
The most popular implementation is based on
the Latent Dirichlet Allocation (LDA) \cite{blei2003}.

According to this,
a \emph{word} is the basic unit of the lowest granularity.
A {\em vocabulary} consists of words and is indexed by $\{1,2,\ldots,V\}$,
and it is fixed and known a-priori.
Now, a \emph{document} is a series of $L$ words, denoted by
$\bm{w} = (w_1, w_2, \ldots, w_L)$,
where the $l^{th}$ word is denoted by $w_l$,
and a \emph{corpus} is a collection of $M$ documents, denoted by 
$D = \{\bm{w_1}, \bm{w_2}, \ldots, \bm{w_M}\}$
Finally, a \emph{topic} is a distribution over a vocabulary.

For each document $\bm{w}$ in $D$, LDA assumes a generative process
 which is initialized by a random distribution over topics; and for each word in $\bm{w}$ we choose a topic from that topics' distribution. 
Although documents share the exact same set of
topics, each topic comes in different proportions inside each document.

While LDA operates solely over series of words, its objective
is to discover the topic structure, which is initially unknown, and is thus considered as latent.
The topics are noted as $\beta_{1:K}$, where $K$ is their number;
each topic $\beta_k$ is a distribution over the vocabulary, where $k \in \{1,\ldots,K\}$;
and $\beta_{kw}$ is the probability of word $w$ existing in topic $k$.
For the $d^{th}$ document the topic proportion of topic $k$ is $\theta_{dk}$.
The topic assignments for the $d^{th}$ document are denoted by $z_d$,
with $z_{dl}$ being the topic assignment for the $l^{th}$ word 
of the $d^{th}$ document. 
Thus, $\beta, \theta$ and $z$ are the latent variables of the model,
with the only observed variable being $w$, where $w_{dl}$
is the $l^{th}$ word in the $d^{th}$ document.
Considering all the documents, the posterior of a topic structure is given by:
\begin{equation*}
p(\beta_{1:K}, \theta_{1:D}, z_{1:D} \mid w_{1:D}) = 
\frac{p(\beta_{1:K}, \theta_{1:D}, z_{1:D}, w_{1:D})}{p(w_{1:D})}
\end{equation*}
However, computing the probability $p(w_{1:D})$ of a document belonging ta a topic structure 
 is intractable~\cite{blei2012}.
Furthermore, LDA introduces priors,
 so that
$\beta_k \thicksim Dirichlet(\eta)$ and $\theta_d \thicksim Dirichlet(\alpha)$.

Though the topic
structure cannot be efficiently computed, 
there are ways to approximate it~\cite{blei2003}.
The two most prominent alternatives for this are  
Markov Chain Monte Carlo (MCMC)
sampling methods
and variational inference~\cite{jordan1999}.
There are also online variants, such as online LDA~\cite{hoffman2010}, where documents arrive in streams
and $\lambda_{1:K}$ is updated according to each document batch.

The work of~\cite{mamakos2018overlapping} was the first to employ PTM for multiagent learning, while existing literature on multiagent learning~\cite{fudenberg1998theory,tuyls2012multiagent}, in both non-cooperative~\cite{hu2003nash,hu1998multiagent,littman1994markov} and cooperative~\cite{DBLP:conf/ijcai/BalcanPZ15,DBLP:conf/atal/ChalkiadakisB04,DBLP:conf/atal/KrausST03,DBLP:conf/atal/KrausST04} game settings, had been largely preoccupied with the study of RL, PAC learning, or simple belief updating algorithms. 
According to it, each agent $i$ (out of $n$) maintains her own model.
For each  coalition $C$
$i \in C$, 
$i$ observes the achieved utility $u_C$.
The contribution $r_{i,C}$ 
of agent $i$ to coalition $C$
 is known to each other agent $j \in C \setminus i$,
 once $C$ is successfully formed.
To be able to feed this information to their
 models,
agents maintain a vocabulary.
The agent vocabulary includes $n$ words,
one for each agent, indicating their contribution,
plus two words for the utility, one representing the gains and the other
representing the losses, summing up to $n+2$ words in total.
Assuming a game round $t$, the agent $i$ interprets 
the coalition configurations of $C, i \in C$, 
as a document by appending in the document the word that indicates the
contribution of agent $j \in C$ \hspace{0.7mm} $r_{j,C}$ times---where 
$r_{j,C} \in \mathbb{N}^+ $ is
the contribution of $j$ to $C$.
Since words are discrete data, $u_C$ cannot be real-valued. To overcome this, we can apply a simple rounding operation to its value.

Another, ``non-conventional'' use of PTM is given in~\cite{georgara2019learning}. In this setting, agents form coalitions that participate in an unknown hedonic game~\cite{aziz2016hedonic}, and can observe only a few game instances.
The document in this case corresponds to agent identifiers in a coalition and a binary label (`gain' or `loss') indicating the sign of the revenue that accrues by this coalition. 
By applying online LDA it is possible to discover ordinal agent collaboration preferences and learn the latent game parameters.

Finally, in~\cite{tripolitakis2017probabilistic}, PTM has been combined with RL to learn user preferences and generate movie recommendations. This approach models items and users as distributions of topics with Dirichlet priors, leveraging crowdsourced corpora e.g. Wikipedia movie synopses. Explicit attributes are substituted by latent features in both items and users distributions and consequently reinforcement learning is applied for recommendation policy adaptation.
In RL, the learning rate is altered according to if the generated recommendations are close to, or far from the average recommendations that the agent has been giving, this way minimizing changes during strong performance and significantly increasing it to escape from low-rated recommendations.

\section{Open Challenges}
\label{sec:challenges}
As stressed early on in this paper, in our view the main challenges for applying multiagent systems that are able to learn and adapt in real-world applications are the following:
\begin{itemize}
\item {\em Non-stationarity of the environments}. As agents learn independently and evolve their strategies, the environment changes in parallel, making the learning process difficult to converge.
\item {\em The need to dynamically balance the trade-off between stability and adaptation}. Stable learning processes are more easy to solve and analyze, while adaptability is required since the behavior of other agents is in general unpredictable.
\item {\em Uncertainty and heterogeneity with respect to the environment, agent beliefs, agent capabilities, and agent internal structures}. Sophisticated solutions should account for probabilistic formulations that are able to capture the inherent uncertainty; and also allow for fitting heterogeneous models, as uniformity is rarely found in practice. 
\begin{itemize}
\item We note that a recent framework put forward by Bakopoulos and Chalkiadakis~\cite{bakopoulos2025a} allows for the principled handling of uncertainty to increase exploration efficiency. This adaptive exploration framework prevents the agent from over-exploring previously explored parts of the environment, instead encouraging exploration of states with higher uncertainty. To create an instance of this framework, a user should select (among others) a definition of uncertainty. In addition, all previous adaptive exploration methods can be viewed as special cases of this framework, when initialized with the appropriate definition of uncertainty.
\item In addition, as already mentioned earlier, we note that a AAAI-2025 publication by Chrysomallis and Chalkiadakis~\cite{chrysomallisAAAI2025heterogeneous} introduces
 a novel, {\em heterogeneous actions} variant (HA-DIIQN) of the DIIQN algorithm, enabling implicit imitation between agents with distinct action spaces, and thus broadening the applicability of deep implicit imitation RL to real-world scenarios with diverse agent characteristics.
\end{itemize}
\item {\em Scalability for large-scale deployments}. The methods should be such, so that a significant increase in the number of agents and their interactions would not render them inapplicable. 
\item {\em Game-theoretic analysis and tractability}.
The convergence of a successful learning process must be guaranteed and the solutions should have desired properties considering agent strategies and incentives. The setting of respective bounds is possible via GT analysis~\cite{yang2020overview}.
\end{itemize}

Focusing solely on the domain of GNNs, the size of the input graphs significantly impacts the computational complexity, and thus can become a major issue. Also, as this is a universal problem in ML, overfitting can affect the trained model's ability to generalize in other, perhaps yet unseen circumstances.
In order to devise effective and applicable GNN solutions, one must keep in mind that {\em (a)} lots of data are required for training; {\em (b)} hyperparameters must be carefully tuned in order to obtain acceptable results; {\em (c)} the interpretation and the explainability of the results are not always straightforward; and {\em (d)} applying GNNs in settings that evolve over time is not always possible.

As regards to DRL, here too the existence of NNs in the design necessitate {\em (a)} large amounts of data for training and {\em (b)} lots of computational resources for often prolonged periods of time. With respect to the RL part, it is important to {\em (c)} have a simulator at hand, since performing RL in the physical world may be expensive or even dangerous, and {\em (d)} to keep in mind that trained DRL models may face difficulties to generalize in face of environment states that had not been observed during training.

For PTM, other, novel applications, different than the conventional formulations for document analysis, can be proven promising and valuable.
Overall, the combination of GNNs, DRL, and PTM, being applied to tackle requirements that are complementary to each other, is a promising direction to consider. As the survey of~\cite{Fathinezhad23} also suggests, combinations of GNN and DRL lead to realistic and effective solutions for real-world problems.

Finally, with respect to the domain of GT, there  is the need to (re-)define equilibrium concepts in a way that enables practical strategic decision-making in complex environments, surpassing the limitations of traditional GT assumptions. By integrating heterogeneous agent beliefs into the computation process, novel equilibrium concepts with bounded rationality and flexible performance guarantees may arise---with an ultimate goal to blur the boundary between cooperative and non-cooperative game theory, and to offer novel solutions applicable across various domains.

\section{Conclusions}
\label{sec:conclusions}
In this survey we focused on {\em (i)} adapting ML techniques currently used for discovering unknown model structures, for the task of strategic opponent modeling; and on {\em (ii)} the potential for intertwining such techniques with tractable and novel GT concepts that do not require strong, restrictive assumptions, such as the CPA and SIH ones, that often do not hold in the real world.

Under this prism, we reviewed methods and approaches in the domains of graph neural networks, deep reinforcement learning, and probabilistic topic modeling.
We presented the most known GNN leargning algorithms variations, and referred to past work that has applied them to multiagent strategic settings.
We distinguished among approaches that consider spatial, spectral, and sequential aspects, as well as those that require assumptions related to common priors or agent self interest.
With respect to DRL, we introduced the required background for tabular and single-agent DRL and delved into multiagent-enabled algorithms. We provided an overview related work and highlighted particular approach characteristics such as decentralization, training philosophy, and inter-agent communication or self interest requirements.
Further, we discussed the theoretical basis of the well-known solution concepts such as the Nash equilibrium and the notion of correlated equilibria, as well as desired game theoretic properties such as fairness and stability.
The field of PTM is also interesting due to its capability to fit unknown underlying distributions, such as the underlying structures governing agent beliefs and types. We presented existing work that utilizes PTM in domains other than document classification, for which this was originally designed for.
Finally, we outlined certain open challenges that exist in this highly challenging research domain.

\begin{credits}
\subsubsection{\ackname} The research described in this paper was carried out within
the framework of the National Recovery and Resilience Plan
Greece 2.0, funded by the European Union - NextGenerationEU (Implementation Body: HFRI. Project name: DEEP-
REBAYES. HFRI Project Number 15430).
We also thank Stefano Albrecht for useful comments on an earlier version of this paper.

\end{credits}

\bibliographystyle{splncs04}
\bibliography{refs}
\end{document}